\documentclass{article}

\PassOptionsToPackage{numbers,compress}{natbib}

\usepackage[eandd, preprint]{neurips_2026}

\usepackage[utf8]{inputenc} % allow utf-8 input
\usepackage[T1]{fontenc}    % use 8-bit T1 fonts
\usepackage{hyperref}       % hyperlinks
\usepackage{url}            % simple URL typesetting
\usepackage{booktabs}       % professional-quality tables
\usepackage{amsfonts}       % blackboard math symbols
\usepackage{nicefrac}       % compact symbols for 1/2, etc.
\usepackage{microtype}      % microtypography
\usepackage{xcolor}         % colors
\usepackage{graphicx}
\usepackage{amsmath}
\usepackage{amssymb}
\usepackage{xurl}
\usepackage{bm}  % boldface math
\usepackage{amsthm}
\usepackage{cleveref}
\newcommand{\ttbar}{{t\bar{t}}}
\newcommand{\mtt}{m_{\ttbar}}
\newcommand{\pTtt}{p_{\mathrm{T}}^{\ttbar}}
\newcommand{\chel}{c_{\mathrm{hel}}}

\newcommand{\dphi}{\Delta\phi(\ell_{\mathrm{hel}},\,\bar{t}_{\ttbar})}
\newcommand{\CRPS}{\mathrm{CRPS}}
\newcommand{\var}{\text{Var}}

\title{Pointwise Metrics Mislead: An Evaluation Protocol for Multimodal Inverse Problems}

\author{%
  Mads~H.~Baattrup\thanks{Corresponding author} \\
  Deutsches~Elektronen-Synchrotron\\
  Hamburg, Germany \\
  \texttt{mads.baattrup@desy.de} \\
  \And
  J{\"o}rn~Bach \\
  Deutsches~Elektronen-Synchrotron \\
  Hamburg, Germany \\
  \texttt{joern.bach@desy.de} \\
  \And
  Laurids~Jeppe \\
  CERN \\
  Geneva, Switzerland \\
  \texttt{laurids.jeppe@cern.ch} \\
  \And
  Finn~Labe \\
  Deutsches~Elektronen-Synchrotron \\
  Hamburg, Germany \\
  \texttt{finn.labe@desy.de} \\
  \And
  Alexander~Grohsjean \\
  University of Hamburg \\
  Hamburg, Germany \\
  \texttt{alexander.grohsjean@uni-hamburg.de} \\
  \And
  Christian~Schwanenberger \\
  Deutsches~Elektronen-Synchrotron \\
  Hamburg, Germany \\
  \texttt{christian.schwanenberger@desy.de} \\
  \And
  Peer~Stelldinger \\
  HAW~Hamburg \\
  Hamburg, Germany \\
  \texttt{peer.stelldinger@haw-hamburg.de} \\
}

\begin{document}

\maketitle

\begin{abstract}
    Evaluation in scientific reconstruction is dominated by pointwise metrics -- RMSE, MAE, per-event resolution -- under the implicit assumption that lower error means better reconstruction. We show that this assumption fails structurally for inverse problems with multimodal posteriors. By the law of total variance, point estimators trained to minimize MSE or MAE produce a marginal spectrum strictly narrower than the truth whenever the posterior has nonzero width. The resulting bias is independent of architecture, training, and dataset size, and it compresses precisely the spectral features -- tails, modes, shapes -- that downstream scientific measurements rely on. We propose a three-part evaluation protocol where each step targets a failure mode the others miss: per-event distributional accuracy via CRPS, population-level marginal accuracy via a spectrum-fidelity diagnostic, and uncertainty trustworthiness via coverage-based calibration. On a synthetic benchmark with an analytic posterior and on a realistic many-to-one inverse problem from particle physics, model rankings reverse between pointwise and distributional metrics, and calibration further separates architectures indistinguishable under CRPS. The evaluation protocol, not the model, determines the scientific conclusion.
\end{abstract}

\section{Introduction and Scope}
\label{sec:introduction}
In scientific reconstruction, evaluation is not a neutral postprocessing step: the metrics we report determine which models advance and which physical conclusions hold. Across particle physics, medical imaging, and geophysics, evaluation is dominated by pointwise resolution metrics (root-mean-squared-error (RMSE), mean-absolute-error (MAE), per-event bias) that ask how close each prediction lies to the truth. We show that this convention fails structurally for under-constrained inverse problems with multimodal posteriors. The optimal predictor under mean-squared-error (MSE) is the conditional expectation \citep{bishop2006pattern}, which for multimodal posteriors can fall between modes in regions of vanishing probability density. Models achieving lower MSE do so by collapsing the posterior more aggressively, producing predictions that are individually ``unphysical'' and collectively distort reconstructed spectra. This is an \emph{evaluation oversight} rather than a model failure. 

We present the following contributions:
\begin{enumerate}
    \item A theoretical argument (\cref{sec:spectral-mismodeling}) showing that any point estimator minimizing MSE or MAE produces a marginal spectrum strictly narrower than the truth whenever the posterior has nonzero variance -- a bias independent of architecture, training objective, and dataset size.
    \item A three-part evaluation protocol -- per-event distributional accuracy, population-level spectrum fidelity, and coverage-based calibration -- that diagnoses the failure modes pointwise metrics miss and applies across regression, mixture, and generative model families on a common scale.
    \item An empirical demonstration on a controlled synthetic benchmark with an analytic posterior and on a realistic many-to-one inverse problem from particle physics, showing that model rankings reverse between pointwise and distributional metrics, and that calibration further separates architectures indistinguishable under the per-event distributional accuracy metric.
\end{enumerate}

\paragraph{Scope}
We consider fully-supervised inverse problems where paired latents $z$ and observations $x$ are available via simulation, and the downstream quantity of interest is the marginal $p(z)$ over a dataset rather than a posterior over global parameters. Ground-truth $z$ is assumed accessible at evaluation time. We do not address joint posterior structure across high-dimensional $z$, domain shift between training and deployment, or unsupervised settings. Our analysis is otherwise domain-agnostic: any inverse problem with non-negligible posterior variance falls within scope.

\section{Background and Related Work}
\label{sec:background-and-related-work}

\paragraph{Fully-Supervised Inverse Problems}
We consider the task of recovering a latent state $z \in \mathbb{R}^n$ from noisy observations $x \in \mathbb{R}^m$ related by a forward model ${x} = G({z}, \xi)$, where $\xi$ represents measurement noise or physical stochasticity -- such a pair, $(z,x)$, is commonly called an \emph{event} or a \emph{sample}. In scientific domains -- particle reconstruction, seismic imaging, medical tomography -- $G$ is a high-fidelity simulator that is analytically intractable to invert. When a large corpus of paired $(z,x)$ samples is available, the task becomes a \emph{fully-supervised inverse problem}: learn a surrogate $G^\dagger$ approximating the posterior $p(z\mid x)$. Such problems are frequently ill-posed, admitting multiple or no exact solutions, either through the stochasticity of $\xi$ or through information loss in ${x}$.

\paragraph{Evaluation in Scientific Reconstruction}
Because scientific reconstruction is typically performed event-by-event, evaluation has traditionally followed the same structure, dominated by RMSE, MAE, or resolution and bias of the per-event error distribution. This convention appears universal: RMSE and MAE are the headline metrics in particle reconstruction \citep{toprecognn, spanet}, sparse-view computed tomography \citep{https://doi.org/10.1002/mp.15489}, image super-resolution benchmarks \citep{generalsupervision-with-DL-regressor}, climate downscaling \citep{10.5555/3304652.3304772}, and geophysical inversion \citep{geophysics-with-DL-regressor} and are routinely adopted as the training objective itself. Lower error is implicitly assumed to mean better reconstruction (also when aggregated) -- true for well-constrained unimodal problems, but not when the conditional posterior has nonzero width (\cref{sec:pointwise-evaluation-failure}). Point predictions are also commonly reported, even when the underlying method produces a full posterior: generative models for gravitational-wave parameter estimation are routinely summarized by medians and 90\% intervals \citep{Shen_2022, PhysRevLett.127.241103}, and normalizing flows for neutrino kinematics are ultimately evaluated via point estimates extracted from the posterior \citep{PhysRevD.109.012005}. This ``generative-to-regressive'' bottleneck persists because the community lacks a structured evaluation protocol that rewards preservation of posterior structure.

\paragraph{Connection to Simulation-Based Inference (SBI)}
Learning the surrogate, $G^\dagger$, is an instance of amortized neural posterior estimation (NPE) \citep{doi:10.1073/pnas.1912789117,NIPS2016_6aca9700}. Our setting differs from canonical SBI in two respects: the latent $z$ is per-event and amortized over $\mathcal{O}(10^4 \text{-} 10^6)$ events, and the scientific quantity of interest is the aggregated marginal $p(z)$ rather than a posterior over global parameters. This motivates an evaluation protocol targeting both per-event accuracy and population-level fidelity.

\paragraph{Proper Scoring Rules, Calibration, and Evaluation Principles}
Evaluation of probabilistic predictions requires metrics that reward \emph{faithful} distributions rather than sharp point predictions. A scoring rule $S(F,z)$ is \emph{proper} if its expectation under the true distribution is minimized by $F$ equal to that distribution, and \emph{strictly proper} if uniquely so \citep{Gneiting01032007}. Propriety distinguishes honest probabilistic evaluation from metrics that can be ``gamed'' by posterior collapse.

The continuous ranked probability score (CRPS) is the standard proper scoring rule for univariate distributional prediction, developed in the probabilistic weather forecasting literature \citep{Gneiting01032007,DecompositionoftheContinuousRankedProbabilityScoreforEnsemblePredictionSystems}. For a predictive cumulative distribution $F$ and an observed value $z$, it is defined as:
\begin{equation}
    \label{eq:crps-integral}
    \CRPS(F,z)=\int_{-\infty}^\infty\left[F(t)-\mathbf{1}(t\geq z)\right]^2\mathrm{d}t,
\end{equation}
where $\mathbf{1}(\cdot)$ denotes the indicator function. When the predictive distribution is represented by $N$ posterior samples $\{\hat{z}^{k}\}_{k=1}^N$, the CRPS can be estimated in $\mathcal{O}(N\log N)$ by sorting \citep{JSSv090i12} and the explicit form is given in \cref{app:crps-estimator}.

Two properties are particularly relevant for this setting. First, CRPS reduces to the MAE when $F=\delta(\hat{z})$ is a Dirac distribution at a point prediction $\hat{z}$, making it directly applicable to both generative and regression models on a common scale. Second, the CRPS is reported per-event and then averaged providing the same granularity as RMSE or MAE so dataset-averaged CRPS can be reported as a drop-in replacement for standard pointwise resolution metrics.

Beyond proper scoring, calibration assesses whether nominal credible regions contain the true value at their stated frequencies \citep{hermans2022trustcrisissimulationbasedinference, talts2020validatingbayesianinferencealgorithms, DBLP:journals/corr/GuoPSW17, DBLP:journals/corr/abs-1807-00263}; we use a conformal variant \citep{10.5555/2074094.2074112,pmlr-v206-wang23n,araz2025fitbitesdustconformal} for finite-sample guarantees and cross-family comparability. SBI benchmarks \citep{pmlr-v130-lueckmann21a} and coverage diagnostics \citep{hermans2022trustcrisissimulationbasedinference,talts2020validatingbayesianinferencealgorithms,pmlr-v202-lemos23a} develop these tools for Bayesian posteriors over global parameters. The aggregated-latent setting we consider -- where the scientific quantity of interest is the marginal $p(z)$ over a dataset -- is not their target, and a population-level spectrum diagnostic is not part of their protocols. InverseBench \citep{inversebench} establishes strong reconstruction baselines using primarily pointwise quality metrics. Multimodal posterior structure is not assessed, and the authors note that systematic evaluation of this remains open. Conformal calibration has been applied to HEP \citep{araz2025fitbitesdustconformal}, but combining it with CRPS and spectrum fidelity for cross-family comparison of supervised reconstruction is, to our knowledge, novel.

\paragraph{Current Methods}
Current approaches to scientific inverse problems fall into three categories, each established across multiple domains. \emph{Analytic solvers} exploit the forward model together with domain-specific physical constraints to recover $z$ from $x$ by solving algebraic or variational constraints. Examples span radio-astronomy image reconstruction \citep{radioastronomy-analytical-method}, top quark reconstruction in physics \citep{sonnenschein-PhysRevD.72.095020}, and classical geophysical inversion \citep{geophysics-with-DL-regressor}. These methods are interpretable but tend to degrade under noise and under-determination. \emph{Deep regression} has become the dominant learning-based approach across the domains above, typically trained to optimize pointwise metrics. Population-level regularizers (e.g., maximum-mean-discrepancy (MMD) penalties \citep{Collaboration:2944724}) augment this with marginal-distribution alignment, but do not address per-event structure. \emph{Generative models} offer a path toward full posterior estimation \citep{ardizzone2019analyzinginverseproblemsinvertible}, with applications including neutrino reconstruction \citep{PhysRevD.109.012005} and gravitational-wave parameter estimation \citep{Shen_2022}.
\section{Limitations of Pointwise Evaluation Metrics}
\label{sec:pointwise-evaluation-failure}
Pointwise metrics are misaligned with the goals of multimodal reconstruction. The arguments below are model-independent, applying to any method that extracts point predictions from a multimodal posterior. We focus on the multimodality-induced failure of point estimation. A complementary Jensen-gap failure mode, arising when downstream observables are nonlinear functions of the latent -- also highlighting that the MSE-optimal predictor in latent space is generally not MSE-optimal once mapped to observables -- is treated in \cref{app:jensen-gap}.

\subsection{The Conditional Mean Pathology}
\label{sec:conditional-mean-pathology}
Consider a generic inverse problem: Given an observation $x$, recover a latent quantity $z$ that is distributed according to the conditional posterior $z\sim p(z\mid x)$. The MSE for a point estimator, $f_\theta(x)$:
\begin{equation}
    \mathcal{L}(\theta)
    =
    \mathbb{E}_{z \sim p(z\mid x)}
    \left[
    \|f_\theta(x) - z\|^2
    \right]
  \label{eq:mse-loss}
\end{equation}
is minimized by the conditional expectation, $\mathbb{E}[z\mid x]$ \citep{bishop2006pattern}, where $f_\theta(x)$ is the $G^\dagger$ surrogate and $||\cdot||$ is the L2-norm. When $p(z\mid x)$ is multimodal, the conditional mean can occupy a region of \emph{zero} posterior density. An instructive illustration is a bimodal symmetric posterior with two Gaussian components with means at $\pm a$ and variance, $\sigma^2$:
\begin{equation}
  p(z \mid x)
  = \tfrac{1}{2}\,\mathcal{N}(z;\,+a,\,\sigma^2)
  + \tfrac{1}{2}\,\mathcal{N}(z;\,-a,\,\sigma^2),
  \qquad a \gg \sigma.
  \label{eq:bimodal}
\end{equation}
The MSE-optimal prediction is $f_\theta(x) = 0$, which has posterior density $p(z=0 \mid x) \approx 0$ whenever the modes are well separated ($a \gg \sigma$). In other words, the ``best'' prediction under the squared error is a value that the posterior assigns essentially zero probability. 

\subsection{Spectral Mismodeling under Aggregation}
\label{sec:spectral-mismodeling}
This per-event pathology has direct population-level consequences. In many scientific applications, the final quantity of interest is not a per-event estimate $f_\theta(x)$ but rather a \emph{marginal distribution} of $z$ across the dataset, which can be referred to as a \emph{spectrum} or an \emph{unfolded distribution}. 

By the law of total variance \citep{grimmett2020probability}:
\begin{equation}
  \var[z]
  \;=\;
  \underbrace{\mathbb{E}\left[\var\left[z \mid x\right]\right]}_{\text{within-posterior variance}}
  \;+\;
  \underbrace{\var\left[\mathbb{E}\left[z \mid x\right]\right]}_{\text{variance of point estimates}},
  \label{eq:total-var}
\end{equation}
from which it follows immediately that
\begin{equation}
  \var\left[\mathbb{E}[z \mid x]\right]
  \;\leq\; \var[z],
  \label{eq:var-inequality}
\end{equation}
with equality only when $\var[z \mid x] = 0$ almost surely, i.e. when the true spectrum has no within-posterior variance. It states that the distribution of conditional means is \emph{always narrower} than the true marginal distribution of $z$ whenever $\var[z \mid x] > 0$.

The compression in \cref{eq:var-inequality} is a \emph{bias, not a variance term}: it does not decrease with dataset size and is inherited by any estimator converging to $\mathbb{E}[z\mid x]$. It has \emph{direct physical consequences}: tails and shapes of marginal distributions encode physical parameters, so their compression directly biases downstream measurements.

The same decomposition explains why population-level regularizers -- e.g., MMD penalties added to an MSE objective (e.g. \citep{Collaboration:2944724}) -- reduce but fail to eliminate the bias. A marginal penalty can reshape the aggregate distribution of point predictions, but it cannot correct their per-event placement; matched marginal variance is then achieved by the wrong assignment of predictions to events, leaving the within-posterior variance term of \cref{eq:total-var} unmodelled.

Finally, the within-posterior variance in \cref{eq:total-var} is largest precisely when multimodality is most pronounced (it includes the spread between modes); an analogous argument applies to MAE-trained models, which target the conditional median.

The practical implication is direct: evaluating models only by pointwise metrics such as RMSE actively rewards suppression of posterior structure and penalizes models that preserve it, biasing reconstructed spectra in ways that do not diminish with more data.

\section{Proposed Three-Step Evaluation Protocol}
\label{sec:three-step-evaluation-protocol}
We propose a three-metric protocol -- per-event distributional accuracy (CRPS), population-level spectral accuracy (spectrum fidelity), and uncertainty calibration (coverage) -- where each metric targets a failure mode the others miss. See \cref{app:evaluation-metrics} for more details.

\subsection{Metric I: CRPS (Per-Event Distributional Accuracy)}
\label{sec:crps}

Our per-event metric is the CRPS (\cref{sec:background-and-related-work}). Three properties make it the right choice for evaluating multimodal inverse problems: \emph{(i)} strict propriety ensures that it is minimized only when the predictive distribution equals the true posterior, so it penalizes posterior collapse rather than rewarding it, \emph{(ii)} its reduction to MAE for point-estimators places generative and regression models on a common footing, enabling fair comparison, and \emph{(iii)} it is per-event and averaged, matching RMSE's granularity while measuring distributional rather than pointwise proximity to the truth. Univariate CRPS suffices when downstream measurements are marginal; the energy score \citep{Gneiting2008} extends the protocol to joint posteriors.

\subsection{Metric II: Spectrum Fidelity (Population-Level Accuracy)}
\label{sec:spectrum-fidelity}
CRPS does not measure the marginal, $p(z)$, across the dataset -- the population-level quantity downstream analyses depend on. To evaluate this, we introduce a population-level diagnostic not targeted by existing SBI benchmarks \citep{pmlr-v130-lueckmann21a}, based on the binned $\chi^2$ statistic. Let $\{n_b^\text{true}\}$ and $\{n_b^\text{pred}\}$ denote the histogram bin counts of the true and predicted values of $z$ over the test dataset. The spectrum fidelity is then defined as the Pearson $\chi^2$ \citep{Pearson1992}:
\begin{equation}
    \label{eq:binned-chi2}
    \chi^2_\text{spec} = \sum_{b=1}^B \frac{\left(n_b^\text{pred} - n_b^\text{true}\right)^2}{n_b^\text{true}},
\end{equation}
where the sum runs over $B$ bins, and assuming Poisson uncertainties with the variance in each bin approximated as $\sigma_b^2\approx n_b^\text{true}$ which is valid when $n_b^\text{true} \gg 1$. For generative models, we construct the predicted histogram by using \emph{one random posterior sample per event}. If the model is well calibrated, the resulting set of samples $\{\hat{z}_i\sim p(z\mid x_i)\}_{i=1}^N$ is marginally distributed as $p(z)$ by construction,\footnote{%
    This follows from $p(z)=\int p(z \mid x)\,p(x)\,\mathrm{d}x$ and requires that the learned posterior equals the true posterior; in practice, $\chi^{2}_{\text{spec}}$ simultaneously tests this assumption.
    }
and the expected $\chi^{2}_{\text{spec}}$ equals $B - 1$ (the number of degrees of freedom). For point-estimators, the single prediction per event is used directly.

We report the binned $\chi^2$-metric as the primary spectrum fidelity diagnostic due to its interpretability and direct connection to goodness-of-fit tests standard in scientific analyses. The Earth Mover's distance \citep{710701} is a natural unbinned alternative.

\subsection{Metric III: Calibration (Uncertainty Trustworthiness)}
\label{sec:coverage}

Neither CRPS nor spectrum fidelity directly assesses whether the \emph{width} of the predicted posterior is trustworthy. A model can obtain a reasonable CRPS by placing probability mass near the truth without being calibrated in its uncertainty estimates. The third axis of our evaluation protocol is therefore a calibration diagnostic: a \emph{coverage curve} that plots empirical coverage, $C(\alpha)$, against nominal confidence level, $\alpha$. A perfectly calibrated model produces a coverage curve tracking the diagonal; undercoverage ($C(\alpha) < \alpha$) indicates posteriors that are too narrow, overcoverage indicates posteriors that are too wide. The area between the curve and the diagonal, $\int_0^1|C(\alpha)-\alpha|\,\mathrm{d}\alpha$, quantifies the overall deviation.

The essential requirement is that calibration \emph{is assessed}, not which specific procedure is used. Some approaches are listed in \cref{app:calibration}. Expected-coverage curves are standard in the SBI literature \citep{hermans2022trustcrisissimulationbasedinference, talts2020validatingbayesianinferencealgorithms, pmlr-v202-lemos23a}. We use a conformal variant because it provides finite-sample guarantees and admits a cross-family comparison that is the point of this protocol \citep{araz2025fitbitesdustconformal}. Following the forecasting principle of ``maximize sharpness subject to calibration'' \citep{RePEc:bla:jorssb:v:69:y:2007:i:2:p:243-268}, well-calibrated models should be ranked by sharpness (e.g., mean prediction-set width).

%Two practical considerations apply regardless of the chosen procedure. First, calibration should also be assessed \emph{conditionally} -- in bins of phase space (e.g. defined by event topology or signal-to-noise ratio) -- to identify regions where uncertainty estimates break down even when global coverage appears acceptable. Second, the choice of nonconformity score (or density estimator) interacts with the training objective. For models trained by maximum likelihood, the negative log-likelihood (NLL) is the natural score. For models trained with likelihood-free objectives, the likelihood may be available but was never a training target; calibration of the resulting density is not guaranteed, even if the density evaluation is correct.
Calibration should also be assessed \emph{conditionally} on phase-space bins (e.g. defined by event topology or signal-to-noise ratio) to surface regional failures hidden by global coverage. The choice of nonconformity score (or density estimator) interacts with the training objective: negative log-likelihood (NLL) is natural for likelihood-trained models but is not guaranteed to be calibrated for likelihood-free objectives, even when density evaluation is correct.

Point estimators without uncertainty cannot be assessed on this axis; this inability is itself a limitation worth reporting given the variance compression of \cref{sec:spectral-mismodeling}.

\section{Benchmark I: A Synthetic Inverse Problem with known Multimodal Posterior}
\label{sec:toy-model}
We first demonstrate the failure of pointwise evaluation in a controlled setting where the true posterior is known analytically.

% === TABLE ===
\begin{table}[t]
\centering
\caption{%
  Synthetic benchmark metrics (10{,}000 test events). Values are mean $\pm$ standard deviation across 5 random seeds. The regression wins on RMSE; the normalizing flow and MDN dominate all distributional metrics. Averaging the normalizing flow posterior recovers the regression's RMSE, CRPS, and spectral mismodeling. Calibration can only be assessed for uncertainty-aware architectures. $\chi^2$ has 49 degrees of freedom; Best per column in \textbf{bold}.}
\label{tab:benchmark-1-results}
\small
\setlength{\tabcolsep}{5pt}
\begin{tabular}{@{}lcccc@{}}
\toprule
Method
  & RMSE\,$\downarrow$
  & CRPS\,$\downarrow$
  & $\chi^{2}_{\text{spec}}$ (closest to 49)
  & Calibration deviance\,$\downarrow$ \\
\midrule
Regression MLP
  & $\mathbf{2.877}^{*}$
  & $2.487$
  & $(4.34 \pm 0.16) \times 10^{5}$
  & --- \\
Heteroscedastic reg.
  & $4.049 \pm 0.037$
  & $1.509 \pm 0.001$
  & $(5.4 \pm 0.1) \times 10^{3}$
  & $\mathbf{0.0057 \pm 0.0001}$ \\
MDN$^{\ddagger}$
  & $4.062 \pm 0.038$
  & $1.286 \pm 0.001$
  & $61.0 \pm 7.0^{\dagger}$
  & $0.0109 \pm 0.0001$ \\
Normalizing flow (NF)
  & $4.091 \pm 0.011$
  & $\mathbf{1.285 \pm 0.002}$
  & $\mathbf{53.2 \pm 11.4}^{\dagger}$
  & $0.0117 \pm 0.0005$ \\
NF (posterior average)
  & ${2.879 \pm 0.002}$
  & $2.486 \pm 0.002$
  & $(1.82 \pm 0.02) \times 10^{5}$
  & --- \\
\bottomrule
\end{tabular}\\[4pt]
{\footnotesize $^{*}$\,Lowest by construction (\cref{sec:conditional-mean-pathology}); converged to constant prediction with negligible seed variance. $^{\dagger}$\,One random posterior sample for $\chi^{2}_{\text{spec}}/\text{ndf}$. $^{\ddagger}$\,Mixture density network needs manual initialization; see \cref{app:benchmark-1-model-architectures-and-training-details}.}
\end{table}

\begin{figure}
    \centering
    \includegraphics[width=\linewidth]{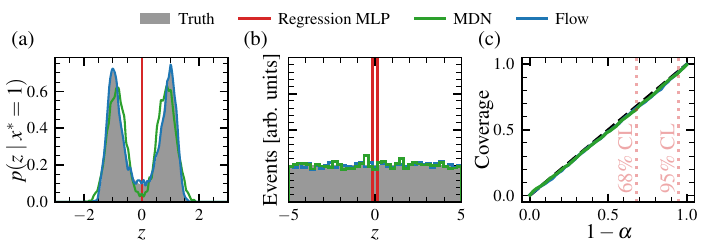}
    \caption{Synthetic benchmark results. \textbf{(a)} Per-event posteriors for an observation ($x^*=1$) in bimodal regime. \textbf{(b)} Marginal distribution of reconstructed $z$ over 10{,}000 test events. \textbf{(c)} Global conformal calibration coverage curves for the flow and MDN models.}
    \label{fig:toy-model}
\end{figure}

\paragraph{Problem design}
We consider the forward model, $f:\mathbb{R}\rightarrow \mathbb{R}$:
\begin{equation}
    \label{eq:benchmark-1-forward-model}
    x=f(z)=z^2+\varepsilon,\qquad \varepsilon\sim \mathcal{N}\left(0, \sigma_\varepsilon^2\right),
\end{equation}
with uniform prior $z\sim\mathcal{U}\left(-a, a\right)$, where $a=5$ and $\sigma_\varepsilon=0.5$. The mapping $z\mapsto z^2$ is symmetric: for any observation $x> 0$, the values $\pm\sqrt{x}$ are equally likely \emph{a priori}, producing an analytically tractable bimodal posterior (see \cref{app:benchmark-1-details}). In this model, $x$ controls a smooth transition from unimodal ($x \approx 0$) to strongly bimodal regimes, covering the generic many-to-one pathology.

\paragraph{Compared methods}
We compare four models on identical training data, differing only in their treatment of per-event uncertainty. The \emph{regression MLP} is a 4-layer MLP trained with MSE; by posterior symmetry, its MSE-optimal prediction is $\hat z(x) = \mathbb{E}[z\mid x] = 0$ for all $x$, providing no per-event uncertainty and cannot be calibrated using conformal prediction. The \emph{heteroscedastic regression} shares the backbone but outputs $(\mu, \sigma)$ trained by Gaussian NLL, modeling per-event uncertainty under a unimodal Gaussian assumption. The \emph{mixture density network} (MDN) shares the regression backbone but outputs the parameters of a $K=2$ Gaussian mixture, trained by NLL; this is our lightweight posterior-preserving baseline. The \emph{conditional normalizing flow} is a 4-layer autoregressive rational-quadratic spline (ARQS) flow \citep{NEURIPS2019_7ac71d43} trained by NLL \citep{papamakarios2021normalizingflowsprobabilisticmodeling}, included as a more expressive reference. More details can be found in \cref{app:benchmark-1-details}.

\paragraph{Evaluation setup}
Following \cref{sec:three-step-evaluation-protocol}, we evaluate the models on 10{,}000 held-out samples. For the uncertainty-aware architectures, we draw $500$ samples per event to approximate the conditional posterior (see \cref{app:benchmark-1-additional-results} for CRPS convergence study). We compute CRPS, $\chi_\text{spec}^2$ with $B=50$ bins over $[-a,a]$, and a conformal calibration where we first split the test set into a calibration set ($n_{\text{cal}} = 1{,}000$) and an evaluation set ($n_{\text{eval}} = 9{,}000$). We use the NLL as nonconformity score, and for this reason, we only assess calibration for the uncertainty-aware models.

\paragraph{Results under pointwise metrics}
\Cref{tab:benchmark-1-results} (first column) confirms that the regression model achieves the lowest RMSE. The normalizing flow posterior mean, which is obtained by averaging all 500 samples per event, recovers approximately the same RMSE, confirming that the regression is simply the conditional expectation of the normalizing flow's posterior. Under RMSE alone, the regression model is the best available method.

\paragraph{Results under the proposed protocol}
The remaining columns of \cref{tab:benchmark-1-results} invert the ranking. The flow and MDN both dominate every distributional metric: CRPS scores are roughly half of the regression's, and $\chi^2_\text{spec}/\text{ndf}\approx 1$ indicates reconstructed marginal spectra statistically consistent with the truth, compared to $\chi^2_\text{spec}/\text{ndf}\sim10^4$ for the regression. \Cref{fig:toy-model} makes the mechanism visible. At $x^*=1$, the true posterior is bimodal at $\pm 1$; the regression predicts $\hat z \approx 0$, a region of low posterior density, while the flow recovers both modes (panel a). Aggregated over the test set, the regression's marginal collapses to a spike at zero and the flow's matches the true uniform on $[-a,\, a]$ (panel b) -- the extreme case of \cref{eq:var-inequality}. The heteroscedastic regression's $\chi^2_\text{spec}$ remains far above the MDN's despite per-event uncertainty -- the unimodal Gaussian family cannot represent the true posterior.

All uncertainty-aware methods achieve nominal coverage (\cref{fig:toy-model}(c), deviance $<0.02$), but with very different sharpness: the heteroscedastic regression's 90\% prediction sets span most of the prior, while the flow's and MDN's are disjoint, concentrated on the modes, and substantially narrower. Coverage alone is therefore insufficient -- see \cref{tab:benchmark-1-results} for marginal deviances and \cref{app:benchmark-1-additional-results} for prediction-set widths and conditional calibration.

\paragraph{Takeaways}
The same models on the same test data yield opposite rankings depending on the metric, but the deeper finding is that the relevant axis is posterior-family expressivity, not generative-vs-regression. The MDN, sharing the regression's backbone with only a 2-component mixture head, matches the flow on every distributional metric; the heteroscedastic regression, with per-event uncertainty but a unimodal Gaussian family, improves over the regression by two orders of magnitude on $\chi^2_\text{spec}$ yet remains far from the flow and MDN because its predictive family cannot represent the true posterior. The flow remains the right default when posterior geometry is unknown a priori, which motivates using flows in Benchmark II.

\section{Benchmark II: A Real-World Multimodal Inverse Problem}
\label{sec:top-reconstruction}

% === TABLE (dphi only) ===
\begin{table}[t]
\centering
\caption{%
  Physics benchmark for the $\dphi$ observable. Five methods evaluated 
  under pointwise (RMSE) and distributional (CRPS, $\chi^{2}_{\text{spec}}$) 
  metrics. The transformer wins on RMSE; the discrete flow dominates 
  distributional metrics. Best per column in \textbf{bold}.}
\label{tab:benchmark-2-results}
\small
\setlength{\tabcolsep}{3.5pt}
\begin{tabular}{@{}l cccc@{}}
\toprule
Method
  & RMSE\,$\downarrow$
  & CRPS\,$\downarrow$
  & $\chi^{2}_{\text{spec}}$/ndf (closest-to-1)
  & Calibration\ deviance\,$\downarrow$\\
\midrule
Analytic
  & $1.740$ & $1.036$ & $462\,/\,99$ & --\\
Transformer (MSE\,+\,MMD)
  & $1.517$ & $0.950$ & $11{,}015\,/\,99$ & -- \\
Transformer (pure MSE)
  & $\mathbf{1.506}$ & $0.943$ & $14{,}901\,/\,99$ & -- \\
Discrete NF
  & $1.780$ & $\mathbf{0.552} $& ${262\,/\,99}$\rlap{$^{\dagger}$} & $\mathbf{0.0238}$ \\
Continuous NF
  & $1.816$ & $0.566$ & $\mathbf{224\,/\,99}$\rlap{$^{\dagger}$} & $0.0886$ \\
\bottomrule
\end{tabular}\\[4pt]
{\footnotesize $^{\dagger}$\,One random posterior sample for $\chi^2_\text{spec}\,/\,\text{ndf}$. Extended table with more observables in \cref{app:extended-results}.}
\end{table}
We benchmark our approach on a particular particle physics inverse problem: reconstructing top-quark pairs in dileptonic decays (see \cref{app:benchmark-2-details}). The problem exhibits all the pathologies the protocol targets: underdetermination (two neutrinos escape detection), combinatorial assignment, non-Gaussian posterior geometry, and measurement noise. Full physics background, observable definitions, and the forward model are in \cref{app:benchmark-2-details}. The latent $z$ encodes the kinematics of two top-quarks; observations $x$ are detector-level (jets, leptons, missing transverse energy). Unlike Benchmark I, no analytic posterior is available; evaluation uses the per-event true latent ${z}$ as reference. Reconstructed spectra feed directly into top-mass measurements and new-particle searches \citep{10.21468/SciPostPhysCore.8.3.053, toponium_discovery}, with parameters of interest encoded in shapes and tails -- exactly the regime \cref{sec:spectral-mismodeling} warns about.

\paragraph{Problem design}
We use a public Delphes \citep{delphes} Monte Carlo dataset \citep{delphes-dataset} (\textasciitilde{}950k simulated events with realistic detector effects) as our forward model. Inputs ${x}$ are detector-level observables (jets, leptons, missing transverse energy); targets ${z}$ are the top-quark four-momenta. Evaluation focuses on a multimodal derived observable $\dphi$; its definition and additional observables are given in \cref{app:physics-observables}.

\paragraph{Compared methods}
We evaluate four methods under identical conditions. The \emph{analytic solver} \citep{sonnenschein-PhysRevD.72.095020} algebraically imposes the kinematic mass constraints of the decay topology to return a single neutrino solution per event. The \emph{transformer regression} (inspired by \citet{Collaboration:2944724}) is a transformer encoder trained with a hybrid MSE\,+\,MMD loss; we also report a pure-MSE variant. The \emph{discrete normalizing flow} (inspired by $\nu^2$-flows \citep{PhysRevD.109.012005}) is a conditional invertible network with exact likelihood. The \emph{continuous normalizing flow} uses the conditional flow-matching objective \citep{chen2018neuralode, lipman2023flowmatchinggenerativemodeling}, with likelihood and sampling via ODE integration. We use a conformal calibration procedure \citep{araz2025fitbitesdustconformal}. Full details in \cref{app:benchmark-2-details}.

\paragraph{Evaluation setup}
We follow the protocol of \cref{sec:three-step-evaluation-protocol}, reporting CRPS, spectrum fidelity, and calibration coverage alongside RMSE for $\dphi$. We draw $500$ posterior samples per event from each flow; CRPS uses the full ensemble, $\chi^2_\text{spec}$ a single random draw per event. Calibration coverage is assessed for the flows via their log-likelihoods; it is unavailable for the point estimators, a limitation of those methods rather than a protocol gap. Results are single-seed per architecture; \cref{app:benchmark-2-evaluation} shows that across four discrete-flow configurations, metric spread is orders of magnitude below the inter-method gaps reported in \cref{tab:benchmark-2-results,tab:extended-benchmark-2-results}.

\begin{figure}
    \centering
    \includegraphics[width=\linewidth]{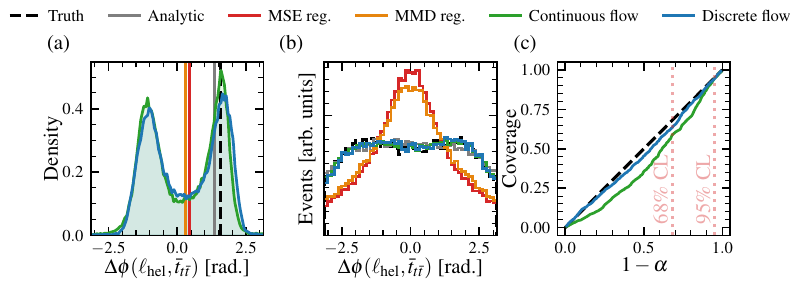}
    \caption{Top reconstruction benchmark results. \textbf{(a)} Reconstructed per-event posterior over $\dphi$. The flow-based posteriors are nearly identical. \textbf{(b)} Marginal distribution of $\dphi$ over the test set. The flows sample a random point per event; the point estimators provide a point estimate without uncertainties. \textbf{(c)} Conformal coverage curve for the flows.}
    \label{fig:top-reconstruction}
\end{figure}

\paragraph{Results under pointwise metrics}
\Cref{tab:benchmark-2-results} shows that the pure-MSE transformer achieves the best RMSE on $\dphi$, with the MMD-regularized variant trailing slightly. The analytic solver's RMSE is comparable to the generative models rather than the transformers, because it commits to a single mode per event rather than computing a posterior mean: when the selected mode is correct, the error is small, when it is not, the error is of order the inter-mode separation. Both the solver and a single generative sample are ``mode-picking'' estimators; the transformers achieve lower RMSE precisely by averaging across modes, which is the source of the spectral distortion identified in \cref{sec:spectral-mismodeling}. The MMD transformer retains competitive RMSE and closes most of the marginal gap on unimodal observables (see \cref{app:benchmark-2-details}), making it the natural choice under current evaluation practice \citep{Collaboration:2944724}. The generative models offer no obvious advantage on either axis.

\paragraph{Results under the proposed protocol}
The distributional metrics flip this conclusion. On $\dphi$, the discrete flow achieves the best CRPS and lowest $\chi^2_\text{spec}$, while the performance of both transformers remains orders of magnitude below that of the flow despite the MMD penalty. This pattern is exactly the prediction of \cref{sec:spectral-mismodeling}: the marginal penalty fixes what marginal information can fix -- on unimodal observables, MMD reduces $\chi^2_\text{spec}$ by an order of magnitude and approaches that of the flows (\cref{app:benchmark-2-details}) -- but on the multimodal $\dphi$ it cannot correct the per-event conditional, and the spectral distortion persists (\cref{fig:top-reconstruction}(b)). Per-event posteriors (\cref{fig:top-reconstruction}(a)) make the mechanism visible: both flows capture the bimodal structure of $\dphi$ while both transformers collapse to the mean between modes, exactly like the synthetic pathology of \cref{fig:toy-model}. The variance compression is visually clearest on unimodal observables (see \cref{app:extended-results}).

The coverage curves (\cref{fig:top-reconstruction}(c)) expose a qualitative difference invisible to RMSE or CRPS: the discrete flow is well-calibrated with only slight undercoverage, while the continuous flow systematically undercovers despite comparable CRPS. The discrepancy possibly reflects both approximate ODE-based density evaluation and the mismatch between the flow-matching objective and the likelihood used for calibration -- the discrete flow optimizes this likelihood directly. Two architectures indistinguishable on per-event accuracy thus produce qualitatively different uncertainty estimates, surfaced only by the full three-metric protocol.

\paragraph{Takeaways}
The physics benchmark confirms and extends the synthetic results. On $\dphi$, model rankings flip between pointwise and distributional evaluation: the pure-MSE transformer wins on RMSE while the discrete flow dominates CRPS and spectrum fidelity. The MMD-regularized transformer partially mitigates spectral distortion on unimodal observables (\cref{app:benchmark-2-details}) but cannot correct the per-event multimodality of $\dphi$, validating the claim of \cref{sec:spectral-mismodeling}. Among the generative models, the discrete and continuous flows appear approximately interchangeable under CRPS and $\chi^2_\text{spec}$, but calibration exposes a consequential architecture difference: exact-likelihood flows produce trustworthy posteriors while approximate-likelihood flows do not. The three-step protocol is necessary to surface these distinctions.

\section{Discussion}
\label{sec:discussion}
\paragraph{Limitations}
%Our CRPS is univariate, computed per observable; for high-dimensional $z$, this loses joint posterior structure, and a multivariate proper score such as the energy score paired with joint-posterior coverage diagnostics such as SBC \citep{talts2020validatingbayesianinferencealgorithms} or TARP \citep{pmlr-v202-lemos23a} would be required to capture it \citep{Gneiting2008}. 
Univariate CRPS loses joint posterior structure for high-dimensional $z$; multivariate proper scores (energy score \citep{Gneiting2008}) and joint coverage diagnostics (SBC \citep{talts2020validatingbayesianinferencealgorithms}, TARP \citep{pmlr-v202-lemos23a}) extend the protocol. The physics benchmark uses a single Monte Carlo generator with Delphes detector simulation, and we have not tested how the learned posteriors or their calibration behave under domain shift to full simulation or real collision data -- validating this robustness is a prerequisite for experimental deployment. Finally, the Benchmark~II models are trained on a modest dataset without exhaustive tuning, so the absolute values in \cref{tab:benchmark-2-results} should not be read as each architecture's best achievable performance. The ranking flip itself is robust: \cref{sec:spectral-mismodeling} shows that no MMD strength can correct per-event multimodality, so while gaps may shrink with tuning, they cannot close.

\paragraph{Applicability beyond the studied benchmarks}
The protocol is domain-agnostic. Any many-to-one inverse problem produces the same posterior multimodality and the same failure of pointwise metrics (e.g. phase-retrieval, cosmological parameter inference, or gravitational wave parameter estimation). The protocol requires no domain-specific infrastructure: CRPS reduces to MAE for point-estimators, spectrum fidelity requires only a binned histogram, and calibration requires either a tractable likelihood or conformal nonconformity scores. The ranking flip is also not limited to bimodal posteriors: by \cref{eq:total-var}, any non-negligible within-posterior variance produces systematic tail compression that does not average away with more data, though the distortion is milder for heavy-tailed unimodal cases and negligible for sharply peaked ones. We recommend practitioners inspect posterior samples for a subset of events before defaulting to RMSE. \footnote{We also note that marginally well-calibrated generative reconstruction is essentially generative unfolding \citep{10.21468/SciPostPhys.18.2.070, 10.21468/SciPostPhys.9.5.074, petitjean2025generativeunfoldingjetssubstructure}.}

\paragraph{Open Problems}
Our findings suggest three directions for future work. First, lightweight per-event uncertainty heads (e.g., MDNs) may correct spectral mismodeling more efficiently than population-level regularizers by directly modeling the (conditional) within-posterior variance of \cref{eq:total-var} \citep{NEURIPS2020_c74c4bf0}. Whether they match flow-based fidelity on realistic multimodal problems is open, and their calibration can be assessed within the same protocol. Second, distributional evaluation gains do not automatically translate into improved measurements: how posterior samples should be carried through downstream inference is application-specific and largely unstudied. Third, closing the calibration gap of likelihood-free generative models likely requires higher-order ODE solvers or likelihood-aware finetuning.

\section{Conclusion}
\label{sec:conclusion}
We have shown that pointwise metrics are structurally misaligned with the scientific goals of supervised reconstruction: the law of total variance forces any point-estimator without per-event uncertainty to produce a marginal spectrum narrower than the truth, a bias that does not diminish with more data. Our three-part protocol -- CRPS, spectrum fidelity, and coverage-based calibration -- addresses each component. We show across a synthetic benchmark with analytic posterior and a realistic many-to-one inverse problem from particle physics that model rankings reverse between pointwise and distributional evaluation, and calibration separates architectures indistinguishable under CRPS. Which model is judged best is a function of how its outputs are scored, not only of what those outputs are. Careful evaluation makes this step visible and exposes where it has been going wrong. By rewarding posterior collapse, pointwise-only evaluation has been favoring models whose reconstructed spectra cannot support the downstream measurements built on them. The protocol we propose makes this visible and gives practitioners a basis for comparison that scientific conclusions can rest on. %Docs, code and evaluation pipelines are released and links can be found in \cref{app:reproducibility}. 
Code and evaluation pipelines are available from the corresponding author on request; see \cref{app:reproducibility}.

\begin{ack}
We thank D.~Stafford for providing a Python implementation of the analytic reconstruction of dileptonic top-quark pairs and for fruitful discussions. The authors acknowledge support from Deutsches Elektronen-Synchrotron (DESY), a member of the Helmholtz Association, and from DASHH, Data Science in Hamburg -- Helmholtz Graduate School for the Structure of Matter, as well as funding by the DFG under Germany's Excellence Strategy -- EXC 2121 ``Quantum Universe'' -- 390833306. This research was supported in part through the Maxwell computational resources operated at DESY, Hamburg, Germany.
\end{ack}

\bibliographystyle{unsrtnat}   % ordered by citation appearance
\bibliography{references}

%%%%%%%%%%%%%%%%%%%%%%%%%%%%%%%%%%%%%%%%%%%%%%%%%%%%%%%%%%%%

\appendix

% Source - https://tex.stackexchange.com/a/121055
% Posted by Andrew Swann, modified by community. See post 'Timeline' for change history
% Retrieved 2026-04-13, License - CC BY-SA 4.0
\crefalias{section}{appendix}
\crefalias{subsection}{appendix}
\crefalias{subsubsection}{appendix}

\section{Evaluation Metrics}
\label{app:evaluation-metrics}

\subsection{Empirical CRPS estimator}
\label{app:crps-estimator}
For a predictive distribution represented by $N$ posterior samples $\{\hat z^{(k)}\}_{k=1}^N$, the CRPS of \cref{eq:crps-integral} is estimated as
\begin{equation}
  \widehat{\mathrm{CRPS}} = \frac{1}{N}\sum_{k=1}^{N}\bigl|\hat{z}^{(k)} - z\bigr|
    - \frac{1}{2N^{2}}\sum_{k=1}^{N}\sum_{j=1}^{N} \bigl|\hat{z}^{(k)} - \hat{z}^{(j)}\bigr|,
  \label{eq:crps-empirical}
\end{equation}
computable in $\mathcal{O}(N \log N)$ via sorting \citep{JSSv090i12}. The estimator has finite-sample bias of $\mathcal{O}(1/N)$ from the second term \citep{https://doi.org/10.1002/met.45}. For the sample sizes used in this work ($N=500$ for both benchmarks) the bias $\approx0.05\%$ of the reported CRPS values is well below the gaps between methods. Reducing $F$ to a Dirac distribution at a point prediction $\hat z$ recovers $\widehat{\mathrm{CRPS}} = |\hat z - z|$, the MAE.

\subsection{Spectrum fidelity}
The binned $\chi^2$ statistic of \cref{eq:binned-chi2} requires only that bin counts are well-approximated as Poisson, i.e.\ $n^\text{true}_b \gtrsim 10$. Bin counts and per-benchmark binning choices are stated in the respective evaluation subsections (\cref{app:benchmark-1-details,app:benchmark-2-details}). For tail-heavy spectra where some bins violate the Poisson approximation, a likelihood-ratio statistic such as Baker-Cousins \citep{BAKER1984437} is a more accurate goodness-of-fit measure. We discuss this where relevant in \cref{app:extended-results}.

\subsection{Calibration}
The conformal procedure used throughout this work, including the choice of nonconformity score per model and the finite-sample marginal coverage guarantee, is described next.

\section{Alternative Calibration Procedures}
\label{app:calibration}
The protocol of \cref{sec:coverage} uses conformal calibration as a concrete instantiation of the calibration axis, but the protocol's requirement is that calibration is assessed -- not which procedure is used. We briefly review three methods and indicate when each is appropriate.

\paragraph{Conformal calibration} Given a nonconformity score $s(z, x)$ (e.g. the NLL), the threshold $\hat q_{1-\alpha}$ is the $\lceil(1-\alpha)(n_\text{cal}+1)\rceil$-th smallest score on a held-out calibration set \citep{araz2025fitbitesdustconformal}. The resulting prediction set satisfies $\mathbb{P}(z \in \mathcal{C}_{1-\alpha}(x)) \geq 1-\alpha$ under exchangeability, giving a finite-sample marginal coverage guarantee that holds across model families. We use it throughout this work for cross-family comparability.

\paragraph{HPD credible intervals} For models with tractable densities, the highest posterior density region $\mathcal{R}_\alpha = \{z : p_\theta(z \mid x) \geq \tau_\alpha\}$ tests the learned density directly without requiring a calibration split. It offers no finite-sample guarantee and is most appropriate when the density is the object of interest itself, e.g.\ for likelihood-trained models being assessed on their own training objective.

\paragraph{PIT / CDF-based diagnostics} The probability integral transform of the true value under the predicted CDF should be uniform under correct calibration \citep{RePEc:bla:jorssb:v:69:y:2007:i:2:p:243-268} and deviations reveal systematic biases. PIT is most informative for univariate predictions and provides a richer diagnostic than scalar coverage, but no formal guarantee.

\paragraph{SBC and TARP} For joint posteriors over high-dimensional latents, simulation-based calibration \citep{talts2020validatingbayesianinferencealgorithms} and TARP \citep{pmlr-v202-lemos23a} extend coverage diagnostics to the joint setting. They are the natural choice when the protocol's univariate CRPS is replaced by the energy score (\cref{sec:crps}) for joint-posterior applications.

\paragraph{Choosing among them} We recommend to use conformal prediction when comparing across model families (its finite-sample guarantee is family-agnostic). Use HPD or PIT when assessing a single model's density quality on its own terms. Use SBC or TARP when the latent of interest is high-dimensional and the joint posterior structure matters.

\section{Benchmark I: Synthetic Inverse Problem}
\label{app:benchmark-1-details}
This appendix provides architecture, training, and evaluation details for the synthetic benchmark of \cref{sec:toy-model}, along with extended results supporting the main-text claims.

\subsection{Forward Model}
\label{app:benchmark-1-forward-model}
We want $p(z \mid x)$ for the model $x = z^2 + \varepsilon$, with $\varepsilon \sim \mathcal{N}(0, \sigma_\varepsilon^2)$ and $z \sim \mathcal{U}(-a, a)$. Here, $\mathcal{N}(\mu, \sigma^2)$ denotes the standard normal distribution with mean $\mu$ and variance $\sigma^2$, and $\mathcal{U}(-a, a)$ denotes the uniform distribution on the range $[-a, a]$. From Bayes' theorem:
\begin{equation}
    p(z \mid x) = \frac{p(x \mid z)\, p(z)}{p(x)}.
\end{equation}
For a given $z$ we have $x \mid z \sim \mathcal{N}(z^2, \sigma_\varepsilon^2)$ which is expanded to
$$
p(x \mid z) = \frac{1}{\sqrt{2\pi}\,\sigma_\varepsilon} \exp\!\left(-\frac{(x - z^2)^2}{2\sigma_\varepsilon^2}\right).
$$
The prior is uniform, so it contributes no $z$ dependence on the likelihood. The unnormalized posterior is:
\begin{equation}
    p(z \mid x) \propto p(x \mid z)\, p(z) \propto \exp\!\left(-\frac{(x - z^2)^2}{2\sigma_\varepsilon^2}\right) \quad \text{for } z \in [-a, a]
\end{equation}
The factors $\frac{1}{\sqrt{2\pi}\sigma_\varepsilon}$ and $\frac{1}{2a}$ are constants w.r.t. $z$ and are absorbed into the normalization. This leads to the unnormalized log-posterior:
\begin{equation}
    \log \tilde{p}(z \mid x) = -\frac{1}{2}\left(\frac{x - z^2}{\sigma_\varepsilon}\right)^2, \quad \text{for } z \in [-a, a]
\end{equation}

$p(x)$ is intractable analytically:
\begin{equation}
    p(x) = \int_{-a}^{a} p(x \mid z)\, p(z)\, \mathrm{d}z = \frac{1}{2a} \int_{-a}^{a} \frac{1}{\sqrt{2\pi}\,\sigma_\varepsilon}\exp\!\left(-\frac{(x - z^2)^2}{2\sigma_\varepsilon^2}\right) \mathrm{d}z,
\end{equation}
but constant with respect to $z$. We can evaluate the unnormalized posterior on a grid and normalize it numerically:
\begin{equation}
    \label{eq:normalized-posterior}
    p(z \mid x) = \frac{\exp(\log {p})}{\int_{-a}^{a} \exp(\log {p})\, \mathrm{d}z}.
\end{equation}
We use \cref{eq:normalized-posterior} as the truth posterior to compare against in Benchmark~I, providing an exact ground-truth posterior for every observation, $x$.

\subsection{Data generation and splits}
\label{app:benchmark-1-data-generation}
We sample $z \sim \mathcal{U}(-a, a)$ with $a = 5$, compute $y = z^2 + \varepsilon$ with $\varepsilon \sim \mathcal{N}(0, \sigma_\varepsilon^2)$ and $\sigma_\varepsilon = 0.5$, and split independently-drawn samples into training (50{,}000), validation (10{,}000), and test (10{,}000) sets. Splits are generated with fixed random seeds; because the prior and noise are sampled fresh for each set, train-test contamination is impossible by construction.

\subsection{Model architectures and training details}
\label{app:benchmark-1-model-architectures-and-training-details}
The model architectures used in Benchmark~I are shown in \cref{tab:benchmark-1-architectures}. Reference implementations of all four models, training scripts, and the exact configuration files used to produce \cref{tab:benchmark-1-results} and \cref{fig:toy-model} are available on request (\cref{app:reproducibility}). All models are trained with AdamW \citep{DBLP:journals/corr/abs-1711-05101}(learning rate $10^{-3}$, batch size $512$) on 50{,}000 samples for $80$ epochs to ensure all models are fully converged. We do not employ early stopping, and we use cosine annealing learning rate scheduling. We have trained on a \texttt{NVIDIA A100 80GB PCIe} GPU. Training the MLP-backbone models takes approximately one minute, while training the normalizing flow takes approximately 7 minutes.
\begin{table}[t]
    \centering
    \caption{Benchmark I model comparison. All non-flow models share a 4-layer MLP backbone and differ only in output head and training objective.}
    \label{tab:benchmark-1-architectures}
    \small
    \begin{tabular}{@{}llllll@{}}
    \toprule
    Model & Backbone & Parameters & Output & Loss/Objective & Posterior \\
    \midrule
    Regression       & 4-layer MLP  &   49{,}921 & $\hat z\approx\mathbb{E}[z\mid x]$                & MSE          & $\delta(\hat z)$ \\
    Heteroscedastic  & 4-layer MLP  &  50{,}050  & $(\mu, \log\sigma)$   & Gaussian NLL & $\mathcal{N}(\mu, \sigma^2)$ \\
    MDN              & 4-layer MLP  &   50{,}566 & $\{\pi_k, \mu_k, \log\sigma_k\}_{k=1}^2$ & Mixture NLL  & $\sum_k \pi_k \mathcal{N}(\mu_k, \sigma_k^2)$ \\
    Normalizing flow & 4-layer ARQS & 74{,}596 & flow parameters           & NLL          & $p_\theta(z \mid x)$ \\
    \bottomrule
    \end{tabular}
\end{table}
\paragraph{Shared backbone}
The regression, heteroscedastic regression, and MDN share a 4-layer fully-connected backbone with $128$ hidden units per layer, ReLU activations, and a single scalar input $x$. Models differ only in the output head and training objective; the normalizing flow uses a different architecture entirely (described below). 

\paragraph{MDN initialization}
On the symmetric posterior of $x = z^2+\varepsilon$, MDNs with default random initialization frequently collapse to an effectively unimodal posterior -- a known property of MDN training on symmetric multimodal targets \citep{bishop1994mixture}. We break the symmetry by initializing the two component means at $\mu_1 = +1$ and $\mu_2 = -1$, with no other changes to architecture, optimizer, or training schedule. This stabilizes training across all 5 seeds reported in \cref{tab:benchmark-1-results}. This instability further motivates using more expressive and stabler flows in Benchmark~II.

\paragraph{Conditional normalizing flow}
We use a conditional normalizing flow built with the \texttt{normflows} library \citep{Stimper2023}. The base distribution is a standard 1D Gaussian (fixed mean and unit variance). The flow stacks 4 transform blocks, each consisting of an autoregressive rational-quadratic spline (ARQS) coupling \citep{NEURIPS2019_7ac71d43} followed by an LU-decomposed linear permutation. Each spline transform has 2 hidden blocks with 64 hidden units conditioning on the scalar observation $x$ (1 context channel). Default \texttt{normflows} hyperparameters are used for spline tail bounds and number of knots.

\subsection{Evaluation details}
\label{app:benchmark-1-evaluation-details}
The regression MLP model reports a point estimate per event, and CRPS reduces to MAE for this model. For the remaining models, CRPS is calculated from $N=500$ posterior samples per event using the sorting-based estimator of \cref{eq:crps-empirical}. The estimator bias from finite $N$ is $\mathcal{O}(1/N) \approx 0.05\%$ \citep{https://doi.org/10.1002/met.45}, well below the gaps reported in \cref{tab:benchmark-1-results}.

With $B=50$ uniform bins over $[-5, 5]$, expected counts under the true uniform prior are $\sim 200$ per bin, validating the Gaussian approximation $\sigma_b^2 \approx n^\text{true}_b$. The regression's collapse to $z\approx 0$ concentrates predictions in a single bin, producing the dominant contribution to $\chi^2_\text{spec}$.

We apply conformal prediction as a model-agnostic calibration diagnostic. For the uncertainty-aware models, the nonconformity score is the NLL $s(z, x) = -\log p_\theta(z \mid x)$. The regression has no notion of nonconformity score that scales with input -- a uniform residual across $x$ produces calibrated coverage trivially via fixed-width intervals.

Given a calibration set of size $n_\text{cal}$, the conformal threshold is then given as the empirical threshold on the calibration set corresponding to some mis-coverage level, $\alpha$ (e.g., $0.05$ for $95\%$ confidence) \citep{araz2025fitbitesdustconformal}:
\begin{equation}
    \hat{q}_{1-\alpha} = \text{the}\;\lceil(1-\alpha)(n_\text{cal}+1)\rceil\text{-th smallest value of}\;\{s_i\}.
\end{equation}
A well-calibrated model should produce a coverage curve that tracks the diagonal.

Calibration alone is not enough -- a model can be perfectly calibrated yet produce sets so wide that they are practically useless. We therefore also report the average Lebesgue size of the conformal prediction set:
$$
\hat{C}_{1-\alpha}(y) \;=\; \{\, z : s(z, y) \le \hat{q}_{1-\alpha} \,\}
$$
on a uniform grid of 1{,}000 points over $[-a, a]$. The reported width is the total Lebesgue measure (sum of interval lengths for disjoint regions).

\subsection{Additional Results}
\label{app:benchmark-1-additional-results}
% === TABLE ===
\begin{table}[t]
\centering
\caption{%
  Expanded version of \cref{tab:benchmark-1-results}. Synthetic benchmark metrics (10{,}000 test events). Values are mean $\pm$ standard deviation across 5 random seeds. The regression wins on RMSE; the normalizing flow and MDN dominate all distributional metrics. Averaging the normalizing flow posterior recovers the regression's RMSE, CRPS, and spectral mismodeling. Calibration only possible for uncertainty-aware architectures. $\chi^2$ has 49 degrees of freedom; Best (mean) per column in \textbf{bold}.}
\label{tab:benchmark-1-additional-results}
\small
\setlength{\tabcolsep}{5pt}
\begin{tabular}{@{}lcccc@{}}
\toprule
Method
  & RMSE\,$\downarrow$
  & CRPS\,$\downarrow$
  & $\chi^{2}_{\text{spec}}$ (closest to 49)
  & Cal. dev.$(\times 10^{-2})$\,$\downarrow$ \\
\midrule
Regression MLP
  & $\mathbf{2.877}^{*}$
  & $2.487$
  & $(4.34 \pm 0.16) \times 10^{5}$
  & --- \\
Regression + MMD
  & $2.958 \pm 0.007$
  & $2.587 \pm 0.002$
  & $(2.830 \pm 0.004) \times 10^{4}$
  & --- \\
Heteroscedastic reg.
  & $4.049 \pm 0.037$
  & $1.509 \pm 0.001$
  & $(5.4 \pm 0.1) \times 10^{3}$
  & $\mathbf{0.0057 \pm 0.0001}$ \\
MDN$^{\ddagger}$
  & $4.062 \pm 0.038$
  & $1.286 \pm 0.001$
  & $61.0 \pm 7.0^{\dagger}$
  & $0.0109 \pm 0.0001$ \\
MDN (posterior mean)
  & $2.879 \pm 0.014$
  & $2.486 \pm 0.001$
  & $(1.88 \pm 0.01) \times 10^{5}$
  & --- \\
Normalizing flow (NF)
  & $4.091 \pm 0.011$
  & $\mathbf{1.285 \pm 0.002}$
  & $\mathbf{53.2 \pm 11.4}^{\dagger}$
  & $0.0117 \pm 0.0005$ \\
NF (posterior average)
  & ${2.879 \pm 0.002}$
  & $2.486 \pm 0.002$
  & $(1.82 \pm 0.02) \times 10^{5}$
  & --- \\
\bottomrule
\end{tabular}\\[4pt]
{\footnotesize $^{*}$\,Lowest by construction (\cref{sec:conditional-mean-pathology}); converged to constant prediction with negligible seed variance. $^{\dagger}$\,One random posterior sample for $\chi^{2}_{\text{spec}}/\text{ndf}$. $^{\ddagger}$\,Mixture density network needs manual initialization.}

\end{table}
\begin{figure}
    \centering
    \includegraphics[width=\linewidth]{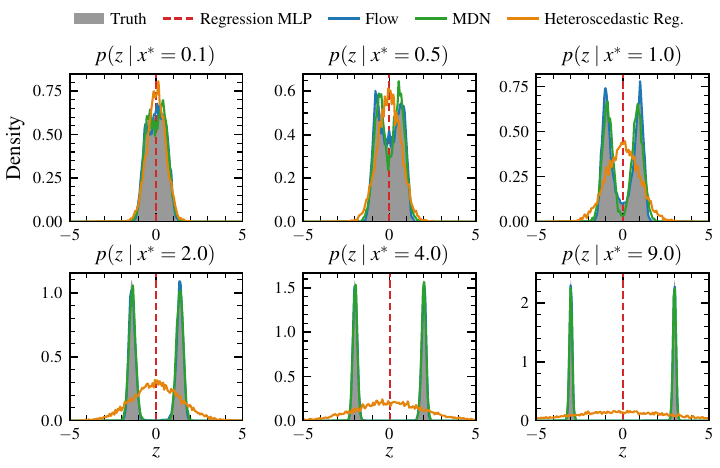}
    \caption{Additional conditional posteriors for the toy model presented in \cref{sec:toy-model}. We present the reconstructed posteriors for six different $x^*$ values similar to \cref{fig:toy-model}(a). Here, we also include the heteroscedastic regression model.}
    \label{fig:toy-model-additional-posteriors}
\end{figure}
\begin{figure}
    \centering
    \includegraphics[width=\linewidth]{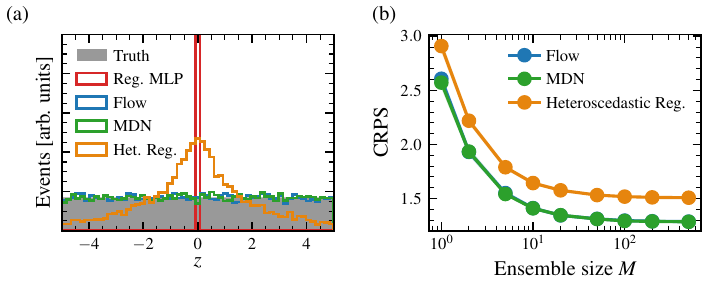}
    \caption{\textbf{(a)} Marginal $z$ over 10{,}000 test events. Similar to \cref{fig:toy-model}(b), but we have included the marginal recovered by the heteroscedastic regression. \textbf{(b)} The sensitivity of the CRPS score is measured by calculating it as a function of ensemble size, $M$, for the distributional models. The flow's curve coincides with MDN's.}
    \label{fig:toy-model-marginal-and-crps}
\end{figure}
\begin{figure}
    \centering
    \includegraphics[width=\linewidth]{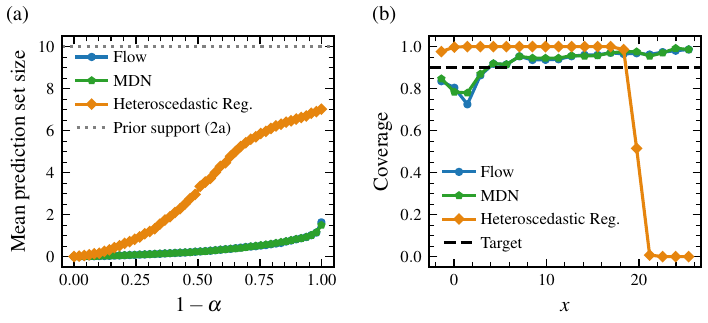}
    \caption{Calibration diagnostics beyond marginal coverage. \textbf{(a)} Mean conformal prediction-set size at each nominal coverage level: all methods are calibrated (\cref{fig:toy-model}(c)) but produce sets of very different widths, demonstrating that coverage and sharpness are independent diagnostics. \textbf{(b)} Empirical coverage at the $90\%$ nominal level as a function of the observation $x$, showing whether conditional calibration is uniform across phase space or hides regional deviations.}
    \label{fig:toy-model-additional-calibration}
\end{figure}

This subsection presents extended visualizations and a sensitivity analysis supporting the main-text claims of \cref{sec:toy-model}.

\paragraph{Per-event posteriors across the unimodal-to-bimodal transition}
\Cref{fig:toy-model-additional-posteriors} extends \cref{fig:toy-model}(a) by showing the recovered per-event posteriors at six different observations $x^*$ spanning the unimodal regime ($x^* \approx 0$, where the two modes merge) through the strongly bimodal regime ($x^* \gg 1$ where modes are well separated at $\pm\sqrt{x^*}$). The flow and MDN track the analytic posterior across all regimes, including the smooth transition between regimes. The heteroscedastic regression -- shown here but not in the main text -- collapses to a single Gaussian centered near zero with inflated variance whenever the true posterior is bimodal, illustrating the failure mode quantified by its $\chi^2_\text{spec}$ in \cref{tab:benchmark-1-results}. Per-event uncertainty is insufficient if the predicted family cannot represent the true posterior geometry.

\paragraph{Marginal recovery including the heteroscedastic regression}
\Cref{fig:toy-model-marginal-and-crps}(a) extends \cref{fig:toy-model}(b) with the heteroscedastic regression's marginal. The flow and MDN marginals match the true uniform on $[-a, a]$ and the regression collapses to a spike at zero. The heteroscedastic regression fall somewhere in between, and it overestimates the center and underestimates the tails. This visually confirms the variance compression argument (\cref{sec:spectral-mismodeling}) that adding per-event uncertainty improves the marginal over a pure point estimator but cannot recover bimodal structure under a unimodal posterior assumption.

\paragraph{CRPS sensitivity to posterior ensemble size}
\Cref{fig:toy-model-marginal-and-crps}(b) shows dataset-averaged CRPS as a function of the number of posterior samples $M$ used for its estimation, for each distributional model. The estimator stabilizes well below the per-method gaps reported in \cref{tab:benchmark-1-results}: by $M \approx 100$ all curves are within Monte Carlo noise of their asymptotic values, justifying the choice of $M = 500$ in the main text and confirming that method rankings are robust to ensemble size. The estimator bias from finite $N$ is $\mathcal{O}(1/N) \approx 0.05\%$ \citep{https://doi.org/10.1002/met.45} when $M=500$.

\paragraph{Prediction-set efficiency}
\Cref{fig:toy-model-additional-calibration}(a) shows the mean prediction-set size as a function of nominal coverage $1-\alpha$ for each model calibrated with conformal prediction. All three uncertainty-aware methods achieve nominal coverage (see \cref{fig:toy-model}(c)), but their prediction sets differ substantially in size. At $90\%$ nominal coverage, the heteroscedastic regression's sets span more than half of the full prior support $[-a, a]$ and convey almost no information beyond the prior. In contrast, the MDN and flow produce sets a few times narrower, concentrated as two disjoint intervals around the true modes $\pm\sqrt{x}$ (\cref{fig:toy-model-additional-posteriors}). This quantifies the claim of \cref{sec:three-step-evaluation-protocol} that coverage alone is an insufficient calibration diagnostic: a model can be perfectly calibrated and yet uninformative, and sharpness must be reported alongside coverage to distinguish and rank the two.

\paragraph{Conditional calibration across the posterior-geometry transition}
\Cref{fig:toy-model-additional-calibration}(b) shows empirical coverage at the $90\%$ nominal level as a function of the observation $x$, probing whether marginal calibration hides regions of systematic over- or undercoverage. This addresses the recommendation of \cref{sec:coverage} that calibration should be assessed conditionally in bins of phase space rather than only marginally. In addition to marginal calibration, we show the conditional calibration in \cref{fig:toy-model-additional-calibration}(b) at $90\%$. The heteroscedastic model achieves marginal calibration by overcovering for small $x$ and undercovering for large $x$. Only conditional calibration makes this visible. The MDN and flow slightly undercover in the unimodal regime and slightly overcover in the strongly bimodal regime.
 
\section{Benchmark II: Dileptonic Top-Quark Reconstruction}
\label{app:benchmark-2-details}
This appendix provides architecture, training, and evaluation details for the physics benchmark of \cref{sec:top-reconstruction}, along with extended results supporting the main-text claims.

\subsection{Physics background}
\label{app:physics-background}

In this work, we evaluate our method on the reconstruction of top quark pairs $\left(t\bar{t}\right)$ produced in proton-proton collisions. The top quark is the heaviest known elementary particle, and its precise reconstruction is a cornerstone of the Large Hadron Collider (LHC) physics program.

\subsubsection{Dileptonic Decay Topology}
Top quarks decay almost exclusively into a $W$ boson and a bottom $\left(b\right)$ quark. When both $W$ bosons from a $\ttbar$ pair decay into a charged lepton $\left(\ell = e, \mu\right)$ and a neutrino $\left(\nu\right)$, the process is called the \textit{dileptonic decay channel}. The full decay chain is represented as:
\begin{equation}
    pp \to t\bar{t} \to \left(W^+ b\right)\big(W^- \bar{b}\big) \to \big(\ell^+ \nu_{\ell} b\big)\big(\ell^- \bar{\nu}_{\ell} \bar{b}\big).
\end{equation}
The experimental signature in the detector consists of two high-momentum leptons, two jets originating from the $b$-quarks ($b$-jets), and significant Missing Transverse Energy $\left(\vec{E}_T^{miss}\;\text{or MET}\right)$ due to the undetected neutrinos.

From an algorithmic perspective, the reconstruction of the parent top quark four-momenta presents three primary challenges:

\begin{enumerate}
    \item \textbf{Underdetermined Kinematics:} While the detector measures the transverse components of the sum of the neutrino momenta $\left(\vec{E}_T^{miss}\right)$, the individual longitudinal momenta $\left(p_z\right)$ and the specific distribution of transverse momentum between the two neutrinos are unknown. This results in a system with six unknown degrees of freedom (the three-momentum components for each neutrino) but only two direct experimental constraints. Even when assuming the known masses of the $W$ boson and the top quark as constraints, the system often remains underdetermined or yields multiple algebraic solutions.
    
    \item \textbf{Combinatorial Ambiguity:} In a standard event, the detector identifies two $b$-jets, but it is not a priori known which jet originated from the top quark and which from the anti-top quark. For $n$ additional light-flavor jets in the event, the number of possible permutations for the final-state assignment grows factorially, creating a complex assignment problem.
    
    \item \textbf{Detector Resolution and Noise:} The measured momenta of jets and the $\vec{E}_T^{miss}$ are subject to experimental uncertainties and resolution effects. Traditional analytical ``kinematic fitting'' methods often fail when the measured values fluctuate such that no physical solution exists for the mass constraints.
\end{enumerate}

\subsection{Dataset details}
\label{app:benchmark-2-dataset-details}
We use the public Delphes \citep{delphes} Monte Carlo dataset of \citet{delphes-dataset} (Zenodo DOI \href{https://doi.org/10.5281/zenodo.8113516}{10.5281/zenodo.8113516}), released under CC-BY 4.0. The dataset comprises $\sim$950k simulated dileptonic $\ttbar$ events with realistic detector effects via Delphes. Event generation, selection, and detector simulation details are in the dataset release and \citet{PhysRevD.109.012005}. We use the dataset's provided train/test splits without modification and we only consider the events generated with MadGraph. 

The dataset was originally constructed for neutrino reconstruction, with bundled targets (called \texttt{neutrinos}) corresponding to the two neutrino four-momenta. For top-quark reconstruction we instead construct the target vectors ourselves from the dataset key \texttt{truth\_particles}. The latent target is two four-vectors per event. The observables per event comprise up to six four-vector jets, two lepton four-vectors, a missing-transverse-energy vector ($p_T$ and $\phi$), and a boolean array indicating whether jets are b-tagged (whether a reconstructed jet is \emph{likely} originating from a bottom quark).

\subsection{Observable definitions}
\label{app:physics-observables}
Once the four-momenta of the top quark $\left(p_t\right)$ and the anti-top quark $\left(p_{\bar{t}}\right)$ are reconstructed and the inverse problem is solved, several high-level observables can be computed for the actual physical study of the process. 
These variables are critical for testing Standard Model predictions and searching for new physics. The distributions over the observables we use here, vary from low-variance and unimodal $\left(\pTtt\right)$ to higher-variance and multimodal $\left(\dphi\right)$.

\subsubsection{Global Kinematic Variables}
These observables describe the overall scale and motion of the $\ttbar$ system.
\begin{itemize}
    \item \textbf{Invariant Mass $\left(\mtt\right)$:} Defined as the Lorentzian norm of the combined four-momentum, $\mtt = \sqrt{(p_t + p_{\bar{t}})^2}$. It represents the total energy scale of the $\ttbar$ production. In the context of machine learning, this is often the most important target for reconstruction, as local deviations in the $\mtt$ spectrum could indicate the existence of heavy resonant particles (e.g., a heavy Higgs boson) decaying into top pairs.
    
    \item \textbf{Transverse Momentum $\left(\pTtt\right)$:} The magnitude of the vector sum of the transverse momenta, $\pTtt = |\vec{p}_{T,t} + \vec{p}_{T,\bar{t}}|$. In a simple leading-order collision, the top quarks are produced back-to-back in the transverse plane, leading to $\pTtt \approx 0$. A non-zero value typically indicates the presence of "Initial State Radiation" (ISR), where the incoming partons radiate gluons before colliding.

\end{itemize}

\subsubsection{Angular and Spin-Sensitive Observables}
These variables describe the relative geometry of the decay products and are sensitive to the quantum state (spin) of the quarks.

\begin{itemize}
    \item \textbf{Azimuthal Angle Difference $\left(\dphi\right)$:} The angle between the lepton and the anti-top quark in the plane transverse to the beam pipe in the helicity frame of the $\ttbar$ system, $\dphi=|\phi_{\ell_\text{hel}}-\phi_{\bar{t}_{\ttbar}}|$. This metric is a close proxy for the spin correlation between the top and the anti-top quarks in a pair system, see the next paragraph for $\chel$.  
    
    \item \textbf{Spin Correlation Variable} $\left(\chel\right)$: The spin correlation observable $\chel$ is computed from the scalar product of the lepton momentum vectors in the rest frames of their respective parent top quarks. It is closely related to $\dphi$ and the equivalent angle between the lepton and its parent top quark. It is a suitable variable to distinguish the spin state of the top quark pair and to search for top quark pairs which were produced by heavier (pseudo-)scalar particles. \\
    To compute $\chel$, one has to boost the lepton momenta into the $\ttbar$ rest frame and then into the parent top quark frames. In this frame, $\chel$ is
    \begin{equation}
        \chel = \hat{\ell^1} \cdot \hat{\ell^2}
    \end{equation}
\end{itemize}

\subsection{Forward model}
\label{app:forward-model}
The forward model maps generator-level top-quark four-momenta $\bm{z}$ to detector-level observations $\bm{x}$ through three stochastic stages: \emph{(i)} hard-scattering matrix-element generation, \emph{(ii)} parton showering and hadronization, and \emph{(iii)} detector response. The \citet{delphes-dataset} dataset uses MadGraph (v3.10) for the hard process, Pythia8 (v8.243) for showering, and Delphes (v3.4.2) for detector simulation. The full configurations are also found in the dataset release \citep{delphes-dataset}.

Two sources of information loss make the inverse problem under-constrained. First, the two neutrinos in dileptonic $t\bar t$ decays escape detection entirely and only their summed transverse component is observable as missing transverse momentum $\left(\vec{E}_T^{miss}\right)$, which provides two constraints in place of the eight neutrino degrees of freedom. Second, parton-jet matching is combinatorial: the two reconstructed bottom-jets can each be associated with either the top or antitop, with no detector-level information distinguishing the two assignments.

Detector effects (energy smearing, angular resolution, jet reconstruction efficiency, bottom-tagging etc.) introduce additional measurement noise on top of the structural under-determination.

\subsection{Model architectures}
\label{app:benchmark-2-model-architectures}
Architectural and training hyperparameters for all four methods are summarized in \cref{tab:benchmark-2-architectures} and \cref{tab:benchmark-2-hyperparameters}. This section provides supplementary detail and design rationale where relevant.

\paragraph{Analytic solver}
We use the algebraic on-shell-mass solver of \citet{sonnenschein-PhysRevD.72.095020}. %as implemented in \href{https://anonymous.4open.science/r/analyticsolver-607E/}{anonymous.4open.science/r/analyticsolver-607E} (MIT License)
Our Python implementation is available from the corresponding author on request.
The solver returns up to 4 discrete neutrino-momentum solutions per event satisfying the on-shell mass constraints. We select the solution per event with $\mtt$ closest to $2m_t$. The solver returns no solution for $\sim 10\%$ of events and these events are excluded from all metrics and spectra for fair comparison (\cref{app:benchmark-2-evaluation}).

\paragraph{Transformer regression}
The transformer regression is inspired by \citet{Collaboration:2944724}, predicting the top and antitop four-momenta directly from the input tokens. The pure-MSE variant uses standard squared-error loss; the MSE\,+\,MMD variant adds two MMD penalties on the $m_{t\bar t}$ and $p_T^{t\bar t}$ marginals (\cref{tab:benchmark-2-hyperparameters}, ``MMD regularization'' block). We attempted MMD penalties on $\chel$ and $\dphi$ across multiple bandwidth and weight configurations and in all cases the MMD term stalled during training without meaningful decrease while MSE continued to improve normally. Marginal regularization requires moving conditional-mean predictions into vanishing per-event posterior density, fighting the MSE objective rather than complementing it when the problem is multimodal. The two losses are structurally incompatible on multimodal observables.

\paragraph{Discrete normalizing flow}
The discrete flow is inspired by the $\nu^2$-flow architecture \citep{PhysRevD.109.012005}: a transformer condition encoder feeds into a stack of conditional ARQS coupling layers with LU-linear permutations. Targets are normalized via an iterative mean-variance layer before flow training, with the corresponding Jacobian included in the likelihood evaluation. The base distribution is standard Gaussian.

\paragraph{Continuous normalizing flow}
The continuous flow uses the conditional flow-matching objective \citep{lipman2023flowmatchinggenerativemodeling}. The vector field $v_\theta(\bm{z}_t, t, \bm{x})$ is parameterized as an encoder-decoder transformer. An encoder processes the conditioning tokens and a decoder cross-attends to the encoded condition while conditioning on the integration time via a Gaussian Fourier embedding (\cref{tab:benchmark-2-hyperparameters}). Sampling and likelihood evaluation use ODE integration and details are listed in the same table. Our transformer adapts baseline-transformer building blocks from the L-GATr codebase \citep{Brehmer:2024yqw,spinner2025lorentz,brehmer2023geometric}.

\subsection{Training details}
\label{app:benchmark-2-training-details}
%%%%%%%%%%%%% Architectures
\begin{table}[t]
    \centering
    \caption{Benchmark~II model architectures. All learned models share the same input representation and training data; they differ in encoder backbone, output head, and training objective.}
    \label{tab:benchmark-2-architectures}
    \footnotesize
    \setlength{\tabcolsep}{4pt}
    \begin{tabular}{@{}lllll@{}}
    \toprule
    Model & Backbone & Output head & Loss & Posterior \\
    \midrule
    Analytic solver & --- & On-shell mass system & --- & up to 4 discrete solutions \\
    Transformer (MSE) & Transformer encoder & MLP $\to \hat{\bm{z}} \in \mathbb{R}^{D_z}$ & MSE & $\delta(\hat{\bm{z}})$ \\
    Transformer (MSE+MMD) & Transformer encoder & MLP $\to \hat{\bm{z}} \in \mathbb{R}^{D_z}$ & MSE + $\lambda\,\text{MMD}$ & $\delta(\hat{\bm{z}})$ \\
    Discrete NF & Transformer encoder & ARQS coupling layers & NLL (exact) & $p_\theta(\bm{z} \mid \bm{x})$ \\
    Continuous NF & Transformer encoder & Conditional vector field & CFM$^{*}$ & $p_\theta(\bm{z} \mid \bm{x})$ (ODE) \\
    \bottomrule
    \end{tabular}\\[4pt]
    {\footnotesize $^{*}$\,Trained with conditional flow matching (CFM) \citep{lipman2023flowmatchinggenerativemodeling}.}
\end{table}

%%%%%%%%%%%%% Hyperparameters
\begin{table}[t]
    \centering
    \caption{Benchmark~II training hyperparameters. Three architectures: a transformer encoder for the MSE/MMD regressions, an encoder-decoder vector field for the continuous flow, and a $\nu^2$-flow-style \citep{PhysRevD.109.012005} architecture combining a transformer condition encoder with a stack of RQS coupling layers for the discrete flow.}
    \label{tab:benchmark-2-hyperparameters}
    \footnotesize
    \setlength{\tabcolsep}{4pt}
    \begin{tabular}{@{}llll@{}}
    \toprule
    & Regression & Continuous flow & Discrete flow \\
    & \footnotesize{(MSE \& MMD)} & \footnotesize{(flow matching)} & \footnotesize{($\nu^2$-flow style)} \\
    \midrule
    \multicolumn{4}{@{}l}{\textit{Condition encoder}} \\
    Encoder blocks       & $8$ & $4$ & $4$ \\
    Attention heads      & $8$ & $8$ & $8$ \\
    Hidden dimension     & $128$ & $128$ & $128$ \\
    Dropout              & $0.1$ & $0.1$ & $0.1$ \\
    Positional encoding dim & $8$ & $8$ & $8$ \\
    \midrule
    \multicolumn{4}{@{}l}{\textit{Flow / decoder head}} \\
    Decoder blocks       & --- & 2 & --- \\
    Time embedding       & --- & Gaussian Fourier (dim 8, scale 30) & --- \\
    Cross-attention      & --- & decoder $\to$ encoded condition & --- \\
    Coupling layers      & --- & --- & 10 RQS stacks \\
    Conditioner MLP      & --- & --- & 2 layers, width 128 \\
    Spline bins / tail bound & --- & --- & 8 / $\pm 3.0$ \\
    Permutations         & --- & --- & LU linear (between stacks) \\
    Initialization       & --- & Gaussian fitted to data & identity \\
    Base distribution    & --- & Gaussian & Gaussian \\
    Target normalization & --- & --- & IterativeNorm (on outputs) \\
    \midrule
    \multicolumn{4}{@{}l}{\textit{Shared input representation}} \\
    Input tokens & \multicolumn{3}{l}{6 jets and 2 leptons (4-momenta) + 1 MET (2D transverse vector); per-token } \\
     & \multicolumn{3}{l}{6-dim PID$^\dagger$ and 1-dim b-tag score when active. Padding handled by attention masks.} \\
    
    positional encoding & \multicolumn{3}{l}{Sinusoidal positional encoding with $d=8$.} \\
    \midrule
    \multicolumn{4}{@{}l}{\textit{Training (all learned methods)}} \\
    Optimizer            & \multicolumn{3}{l}{AdamW ($0.9, 0.999$)} \\
    Learning rate        & \multicolumn{3}{l}{$10^{-3}$ (cosine schedule)} \\
    Batch size           & \multicolumn{3}{l}{$1024$} \\
    Early stopping & \multicolumn{3}{l}{Patience $P=15$ on validation loss evaluated every 1500th batch} \\
    Weight decay         & \multicolumn{3}{l}{$10^{-5}$} \\
    Hardware             & \multicolumn{3}{l}{\texttt{NVIDIA A100 80GB PCIe}} \\
    Training time per model & $\sim 4\,\text{h}$ & $\sim 8\,\text{h}$ & $\sim 16\,\text{h}$ \\
    \midrule
    \multicolumn{4}{@{}l}{\textit{MMD regularization (MMD-regularized transformer only)}} \\
    Total loss & \multicolumn{3}{l}{$\mathcal{L} = \mathcal{L}_\text{MSE} + \lambda_{\mtt} \cdot \mathrm{MMD}^2_{m_{t\bar t}} + \lambda_{p_T} \cdot \mathrm{MMD}^2_{p_T^{t\bar t}}$} \\
    Penalty weights & \multicolumn{3}{l}{$\lambda_{\mtt} = 2 \times 10^4$, $\lambda_{p_T} = 2 \times 10^3$} \\
    Kernel & \multicolumn{3}{l}{Multi-scale RBF, biased estimator} \\
    Bandwidths $\left(\mtt\right)$ & \multicolumn{3}{l}{$\{2,\,20,\,100,\,200,\,400,\,2000\}$ GeV} \\
    Bandwidths $\left(\pTtt\right)$) & \multicolumn{3}{l}{$\{0.8,\,8,\,40,\,80,\,160,\,800\}$ GeV} \\
    Targeted observables & \multicolumn{3}{l}{$m_{t\bar t}$ and $p_T^{t\bar t}$ only; $c_\text{hel}$ and $\dphi$ are not regularized$^*$} \\
    \midrule
    \multicolumn{4}{@{}l}{\textit{Continuous flow}} \\
    ODE solver & \multicolumn{3}{l}{Runge-Kutta 4, rel/abs tolerance $10^{-7}/10^{-9}$} \\
    Integration steps & \multicolumn{3}{l}{$50$} \\
    Hutchinson's samples & \multicolumn{3}{l}{1} \\
    \midrule
    \multicolumn{4}{@{}l}{\textit{Sampling}} \\
    Sampling time per event & --- & $\sim 2.5\,\text{ms}$ without NLL computation & $\sim 0.5\,\text{ms}$ \\
    \bottomrule
    \end{tabular}\\[4pt]
    {\footnotesize $^{*}$\,MMD penalties on $\chel$ and $\dphi$ were attempted but did not converge (see \cref{app:benchmark-2-model-architectures});\,$^{\dagger}$PID is a one-hot-style 6D particle ID encoding of charge and particle type.}
\end{table}

The models used in Benchmark~II are shown in table \cref{tab:benchmark-2-architectures}. Reference implementations of all four models, training scripts, and the exact configuration files used to produce \cref{tab:benchmark-2-results} and \cref{fig:top-reconstruction} are available on request (\cref{app:reproducibility}). All models are trained with AdamW \citep{DBLP:journals/corr/abs-1711-05101} with hyperparameters shown in \cref{tab:benchmark-2-hyperparameters}.

\subsection{Evaluation details}
\label{app:benchmark-2-evaluation}
The analytic solver returns a solution for $\sim90\%$ of events; we restrict marginal-spectrum and metric computation to its solvable subset for fair comparison. The regression and flow-based models reconstruct all events ($100\%$ efficiency), and their metrics on the full test set differ by only a small amount. For the flows we draw $N=500$ posterior samples per event: CRPS uses the full ensemble; $\chi^2_\text{spec}$ uses one random draw per event. Histograms use $B=100$ bins (only bins with nonzero expected content enter in computation) for calculating $\chi^2_\text{spec}$. Conformal calibration uses $n_\text{cal}=1{,}000$ events from the test set, with the remaining events for evaluation. We use the per-event NLL as nonconformity score for both flows.

\paragraph{CRPS sensitivity to posterior ensemble size}
\begin{figure}
    \centering
    \includegraphics[width=0.5\linewidth]{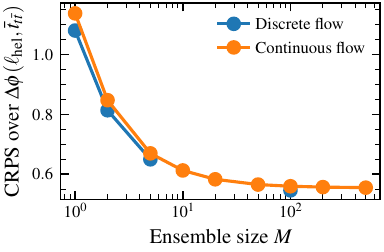}
    \caption{The sensitivity of the CRPS score over $\dphi$ is measured by calculating it as a function of ensemble size, $M$, for the flow-based models.}
    \label{fig:top-reco-crps}
\end{figure}
Based on the discussion in \cref{app:benchmark-1-additional-results} we also include a plot showing dataset-averaged CRPS as a function of the number of posterior samples $M$ used for its estimation in \cref{fig:top-reco-crps}. The estimator stabilizes well below the per-method gaps reported in \cref{tab:benchmark-2-results}: by $M \approx 100$ all curves are within Monte Carlo noise of their asymptotic values, justifying the choice of $M = 500$ in the main text and confirming that method rankings are robust to ensemble size. The estimator bias from finite $N$ is $\mathcal{O}(1/N) \approx 0.05\%$ \citep{https://doi.org/10.1002/met.45} when $M=500$.

\paragraph{Robustness across flow configurations}
\begin{table}[t]
    \centering
    \caption{Robustness across four discrete-flow implementations sharing architecture and hyperparameters but differing in likelihood implementation. Values are mean $\pm$ standard deviation on the test set. Calibration is omitted because it depends on the absolute log-likelihood, which is not comparable across implementations.}
    \label{tab:benchmark-2-flow-robustness}
    \footnotesize
    \setlength{\tabcolsep}{6pt}
    \begin{tabular}{@{}lccc@{}}
    \toprule
    Observable & RMSE & CRPS & $\chi^2_{\mathrm{spec}} / \mathrm{ndf}$ \\
    \midrule
    $p_T^{t\bar{t}}$ [GeV] & $40.58 \pm 1.90$ & $12.99 \pm 0.20$ & $827 \pm 155\,/\,85$ \\
    $m_{t\bar{t}}$ [GeV] & $142.66 \pm 13.51$ & $39.12 \pm 0.19$ & $755 \pm 173\,/\,72$ \\
    $c_{\mathrm{hel}}$ & $0.453 \pm 0.003$ & $0.169 \pm 0.001$ & $228 \pm 29\,/\,99$ \\
    $\Delta\phi(\ell_{\mathrm{hel}}, \bar{t}_{t\bar{t}})$ [rad] & $1.756 \pm 0.004$ & $0.543 \pm 0.004$ & $272 \pm 26\,/\,99$ \\
    \bottomrule
    \end{tabular}
\end{table}
Compute cost precluded multi-seed training per architecture. As an alternative, we trained four discrete-flow variants sharing the same architecture and hyperparameters but differing in how the change-of-variables Jacobian is computed. Because absolute log-likelihoods are not comparable across these implementations, we restrict the robustness check to implementation-invariant metrics: RMSE, CRPS, and $\chi^2_\text{spec}$ (\cref{tab:benchmark-2-flow-robustness}). On CRPS and $\chi^2_\text{spec}$, the uncertainties are two to three orders of magnitude smaller than the inter-method gaps to the transformer baselines in \cref{tab:extended-benchmark-2-results}: for $\dphi$ the CRPS spread is roughly $1\%$ and the $\chi^2_\text{spec}$ spread $0.2\%$ of the corresponding gaps. RMSE uncertainties are larger in relative terms and reach the same order as some inter-method gaps. We interpret this as bounding implementation noise on the metrics that drive the ranking flip of \cref{sec:top-reconstruction} well below the effect sizes themselves. This is a different constraint from seed variance for a fixed implementation: a multi-seed study remains future work.

\subsection{Results across all four observables}
\label{app:extended-results}
% === TABLE ===
\begin{table}[t]
    \centering
    \caption{%
      Benchmark~II: methods evaluated under pointwise (RMSE) and distributional (CRPS, $\chi^{2}_{\text{spec}}$) metrics across four observables (ordered by increasing per-event posterior multimodality). Averaging the flow posterior \emph{in observable space} per event recovers RMSE substantially better than any latent-space regression, demonstrating the Jensen-gap argument of \cref{app:jensen-gap}: the MSE-optimal observable estimator is the conditional mean of the observable, not the observable of the conditional mean. The $\chi^2$ for $\pTtt$ and $\mtt$ is biased since the Gaussian approximation breaks down in the heavy tails (Baker-Cousins \citep{BAKER1984437} would be more appropriate, but within-method comparisons remain valid). Best per column in \textbf{bold}}
    \label{tab:extended-benchmark-2-results}
    \footnotesize
    \setlength{\tabcolsep}{3.5pt}
    \begin{tabular}{@{}l ccc ccc@{}}
    \toprule
    Method
      & RMSE\,$\downarrow$
      & CRPS\,$\downarrow$
      & $\chi^{2}_\text{spec}/\text{ndf}\!\downarrow$
      & RMSE\,$\downarrow$
      & CRPS\,$\downarrow$
      & $\chi^{2}_\text{spec}/\text{ndf}\!\downarrow$ \\
    \midrule
    & \multicolumn{3}{c}{$\pTtt$}
    & \multicolumn{3}{c}{$\mtt$} \\
    \cmidrule(lr){2-4}\cmidrule(lr){5-7}
    Analytic solver
      & $34.20$ & $20.97$ & $1{,}450\,/\,82$
      & $116.57$ & $71.55$ & $4{,}183\,/\,65$ \\
    Transformer (pure MSE)
      & $37.02$ & $25.33$ & $7{,}189\,/\,82$
      & $95.16$ & $69.37$ & $19{,}588\,/\,64$ \\
    Transformer (MSE+MMD)
      & $34.82$ & $24.28$ & ${651\,/\,82}$
      & $97.64$ & $66.76$ & $6{,}778\,/\,64$ \\
    Discrete flow
      & $38.59$ & $\mathbf{12.57}$ & $\mathbf{489\,/\,52}$
      & $148.17$ & $\mathbf{38.39}$ & $\mathbf{815\,/\,38}$ \\
    Discrete flow (post.\ mean)
      & $\mathbf{26.00}$ & $17.48$ & $11{,}430\,/\,82$
      & $\mathbf{82.31}$ & $53.32$ & $9{,}731\,/\,68$ \\
    Continuous flow
      & $47.80$ & $15.14$ & $2{,}165\,/\,78$
      & $131.54$ & $39.97$ & $1{,}569\,/\,68$ \\
    Continuous flow (post.\ mean)
      & $30.61$ & $22.84$ & $23{,}653\,/\,82$
      & $84.34$ & $54.37$ & $9{,}418\,/\,68$ \\
    \midrule
    & \multicolumn{3}{c}{$\chel$}
    & \multicolumn{3}{c}{$\dphi$} \\
    \cmidrule(lr){2-4}\cmidrule(lr){5-7}
    Analytic solver
      & $0.412$ & $0.303$ & $416\,/\,99$
      & $1.740$ & $1.036$ & $462\,/\,99$ \\
    Transformer (pure MSE)
      & $0.402$ & $0.299$ & $2{,}315\,/\,99$
      & $1.506$ & $0.943$ & $14{,}901\,/\,99$ \\
    Transformer (MSE+MMD)
      & $0.421$ & $0.316$ & $3{,}621\,/\,99$
      & $1.517$ & $0.950$ & $11{,}015\,/\,99$ \\
    Discrete flow
      & $0.455$ & $\mathbf{0.168}$ & $297\,/\,99$
      & $1.780$ & $\mathbf{0.552}$ & ${262\,/\,99}$ \\
    Discrete flow (post.\ mean)
      & $0.323$ & $0.245$ & $11{,}196\,/\,99$
      & $\mathbf{1.264}$ & $0.870$ & $31{,}950\,/\,99$ \\
    Continuous flow
      & $0.483$ & $0.174$ & $\mathbf{266\,/\,99}$
      & $1.816$ & $0.566$ & $\mathbf{224\,/\,99}$ \\
    Continuous flow (post.\ mean)
      & $\mathbf{0.331}$ & $0.257$ & $15{,}092\,/\,99$
      & $1.277$ & $0.902$ & $35{,}801\,/\,99$ \\
    \bottomrule
    \end{tabular}
\end{table}

\begin{figure}
    \centering
    \includegraphics[width=\linewidth]{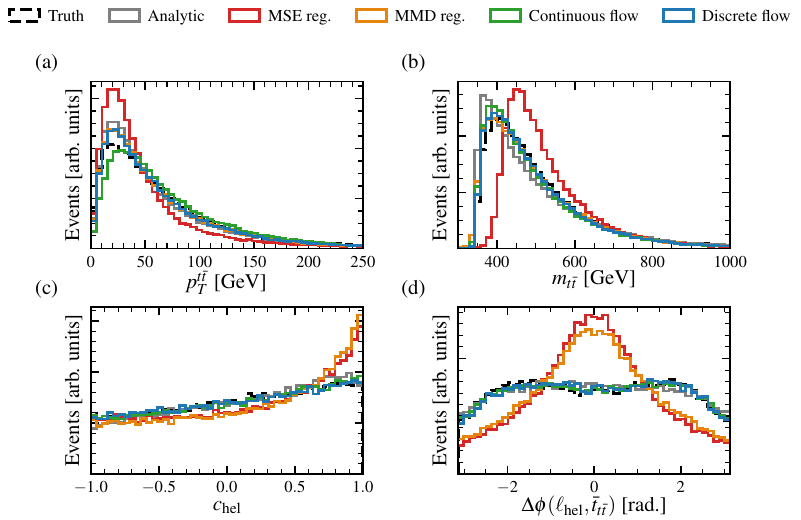}
    \caption{Marginal reconstructed spectra for $\pTtt$, $\mtt$, $\chel$, and $\dphi$ (in order of increasing multimodality). The dashed line shows the truth marginal and the colored curves show each method's reconstructed marginal. Quantitative comparison via $\chi^2_\text{spec}$ is in \cref{tab:extended-benchmark-2-results}.}
    \label{fig:top-reco-additional-marginal-spectra}
\end{figure}
The main text focuses on $\dphi$ as a strongly multimodal observable where our evaluation protocol has the highest impact. This section presents the full evaluation across all four observables ($\pTtt$, $\mtt$, $\chel$, $\dphi$, ordered by increasing multimodality of the per-event posterior). \Cref{tab:extended-benchmark-2-results} shows that the ranking flip described in \cref{sec:top-reconstruction} is consistent across all observables: the pure-MSE transformer wins or near-wins RMSE everywhere disregarding posterior means (see \cref{app:jensen-gap}), while the flows dominate CRPS and $\chi^2_\text{spec}$. 

\paragraph{The MMD penalty pattern}
The MMD-regularized transformer reveals the mechanism predicted in \cref{sec:spectral-mismodeling}. On $\pTtt$, where the per-event posterior is predominantly unimodal, the MMD penalty achieves $\chi^2_\text{spec}/\text{ndf}$, comparable to the flows. On the multimodal $\dphi$, the same penalty produces $\chi^2_\text{spec}/\text{ndf}$ in the thousands. This asymmetry is the empirical signature of the law of total variance: marginal regularization can compensate global distortion when within-posterior variance is small, but cannot correct per-event multimodality when within-posterior variance is large. The intermediate observables ($\mtt$, $\chel$) show intermediate behavior, with MMD reducing but not closing the gap to the flows.

\paragraph{Marginal spectra}
\Cref{fig:top-reco-additional-marginal-spectra} visualizes the same conclusion. The MSE transformer compresses the marginal of every observable, with the distortion most visually severe for $\dphi$. The MMD transformer recovers the unimodal $\pTtt$ and $\mtt$ marginals, where it was specifically tuned (\cref{app:benchmark-2-model-architectures}), deviates noticeably for $\chel$, and fails on $\dphi$. Both flows track all four marginals closely, with no observable-specific tuning.

\paragraph{Variance compression of spectra} The variance compression of \cref{sec:spectral-mismodeling} is visually clearest on $\pTtt$ in \cref{fig:top-reco-additional-marginal-spectra}(a): the MSE transformer's marginal is sharper at the peak and lower in the tail than the truth, with the peak itself shifted upward because the right-tailed underlying distribution pulls conditional means toward higher $\pTtt$ values. The same pattern is also more pronounced in $\mtt$ in \cref{fig:top-reco-additional-marginal-spectra}(b).

\paragraph{Caveat on $\chi^2_\text{spec}$ for heavy-tailed observables}
The $\chi^2_\text{spec}$ values for $\mtt$ and $\pTtt$ reported in \cref{tab:extended-benchmark-2-results} are systematically biased. Both observables have heavy tails with bins where the Poisson approximation $\sigma_b^2 \approx n_b^\text{true}$ underestimates the true uncertainty, inflating $\chi^2$. A proper goodness-of-fit assessment for tail-heavy spectra uses the Baker-Cousins likelihood-ratio statistic \citep{BAKER1984437}, which we leave for future work. Within-method comparisons remain valid because the bias affects all methods equally on a given observable.

\subsubsection{Jensen gap between latent and observable estimators}
\label{app:jensen-gap}
The conditional-mean pathology of \cref{sec:conditional-mean-pathology} has a basis-dependent refinement when downstream quantities are nonlinear functions of the latent. Let $\mathcal{O}: \mathbb{R}^{D_z} \to \mathbb{R}$ be a physics observable computed from the latent four-vector $\bm{z}$. The MSE-optimal estimator for $\mathcal{O}(\bm{z})$ given observations $\bm{x}$ is 
\begin{equation}
    \hat{\mathcal{O}}^* (\bm{x}) = \mathbb{E}[\mathcal{O}(\bm{z}) \mid \bm{x}] = \int \mathcal{O}(\bm{z}) \, p(\bm{z} \mid \bm{x}) \, d\bm{z}.
\end{equation}
A regression trained on the latent under squared-error loss instead converges to $\bm{\hat z}(\bm{x}) = \mathbb{E}[\bm{z} \mid \bm{x}]$ and yields the prediction $\mathcal{O}(\mathbb{E}[\bm{z} \mid \bm{x}])$ for the observable. By Jensen's inequality,
\begin{equation}
    \mathcal{O}(\mathbb{E}[\bm{z} \mid \bm{x}]) \neq \mathbb{E}[\mathcal{O}(\bm{z}) \mid \bm{x}]
\end{equation}
in general, with the gap vanishing only when $\mathcal{O}$ is linear or the posterior $p(\bm{z} \mid \bm{x})$ is degenerate. The regression is therefore not MSE-optimal for the observable even setting aside the multimodality issues of \cref{sec:spectral-mismodeling}.

Generative methods, by contrast, recover the MSE-optimal observable estimator at evaluation time: given posterior samples $\bm{z}^{(k)} \sim p(\bm{z} \mid \bm{x})$, the Monte Carlo average $\frac{1}{N}\sum_k \mathcal{O}(\bm{z}^{(k)})$ converges to $\mathbb{E}[\mathcal{O}(\bm{z}) \mid \bm{x}]$. This explains the systematic RMSE advantage of the ``posterior mean'' rows in \cref{tab:extended-benchmark-2-results}. Across all four observables, both flow posterior means achieve RMSE substantially below all latent-space regression baselines, despite the regressions being trained directly to minimize four-vector MSE. The Jensen gap for these observables is large because top-quark posteriors are wide and the observable maps are strongly nonlinear (squared invariant masses, transverse-plane angles).

This argument is independent of the multimodality argument of \cref{sec:spectral-mismodeling}: even for unimodal posteriors with substantial variance, the Jensen gap persists. It provides a second, independent reason to prefer methods that preserve the per-event posterior over those that collapse it to a four-vector point estimate. The argument also highlights why the basis choice for any latent state regression directly influences the reconstruction RMSE on observables.

\subsection{Per-event posterior structure across observables}
\begin{figure}
    \centering
    \includegraphics[width=\linewidth]{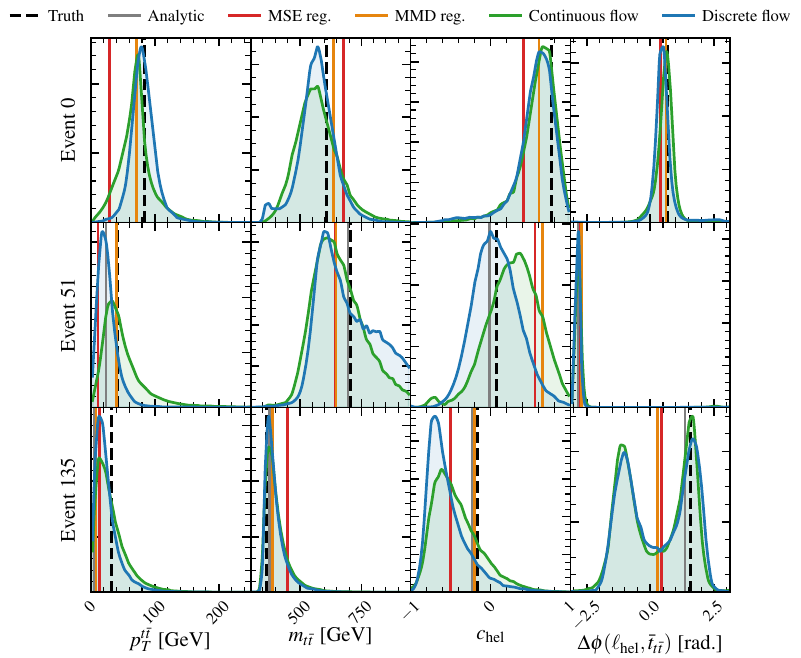}
    \caption{Reconstructed posteriors (filled curves) and point estimates (vertical lines) for three randomly selected test events across the four observables $\pTtt$, $\mtt$, $\chel$, and $\dphi$. True targets are shown as dashed black lines. Event 135 (bottom row of $\dphi$ column) is the event displayed in \cref{fig:top-reconstruction}(a).}
    \label{fig:top-reco-additional-posteriors}
\end{figure}
\Cref{fig:top-reco-additional-posteriors} extends \cref{fig:top-reconstruction}(a) by showing per-event reconstructed posteriors for three randomly selected test events across all four observables. The same per-event collapse pattern observed for $\dphi$ in the main text persists across all observables: both transformer regressions concentrate near the per-event posterior mean, while the flows recover the underlying multimodal or asymmetric structure where present. The contrast is most visually striking on $\dphi$ (strongly multimodal). On $\pTtt$ and $\mtt$ the posteriors are predominantly unimodal but skewed, and the regression's bias is correspondingly milder. Event 135 (bottom-right in the $\dphi$ row) is the same event shown in \cref{fig:top-reconstruction}(a). Across all events, the analytic solver returns at most a single point estimate per event and fails to return any solution for Event 0.

\subsection{Conditional calibration reveals architecture and data limitations}
\begin{figure}
    \centering
    \includegraphics[width=\linewidth]{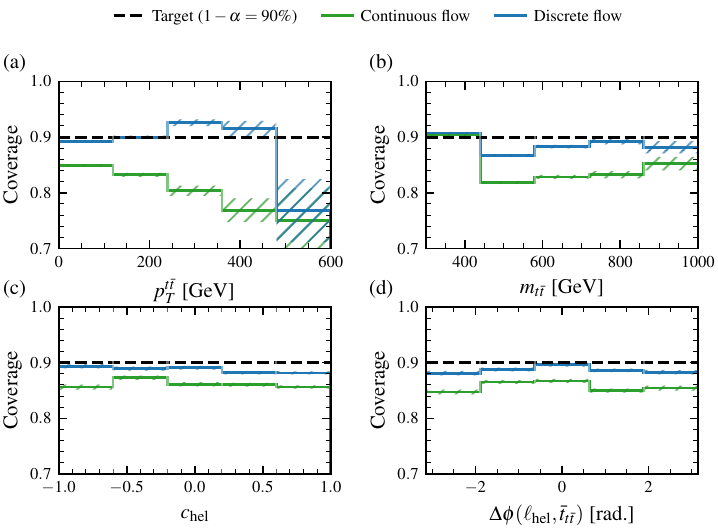}
    \caption{Empirical coverage at the $90\%$ nominal level as a function of the true value of each observable. Shaded bands show binomial uncertainty per bin. The rightmost bins for $\pTtt$ and $\mtt$ have low statistics as indicated by the uncertainty bands.}
    \label{fig:top-reco-conditional-calibration}
\end{figure}
\Cref{sec:coverage} recommends assessing calibration conditionally -- in bins of phase space -- to identify regions where uncertainty estimates fail despite acceptable global coverage. \Cref{fig:top-reco-conditional-calibration} applies this diagnostic to the two flows, plotting empirical coverage at the $90\%$ nominal level as a function of each observable. Both flows undercover throughout, but the discrete flow tracks the nominal level substantially more closely than the continuous flow across all four observables, consistent with the marginal coverage curves of \cref{fig:top-reconstruction}(c). The continuous flow's undercoverage is approximately uniform in observable value, suggesting a global density-evaluation issue (likely from approximate ODE-based likelihood; see \cref{sec:top-reconstruction}) rather than a phase-space-specific failure. The exception is the high-$\pTtt$ region ($\pTtt \gtrsim 450\, \text{GeV}$), where both flows undercover more severely. This regime is sparsely populated in the training data, and the degraded calibration likely reflects extrapolation rather than a model failure. Conditional calibration thus surfaces both a global architectural difference (continuous vs discrete) and a phase-space-specific data limitation, neither of which is visible in the global coverage curve alone.

\section{Reproducibility}
\label{app:reproducibility}
This appendix documents the artifacts released with the paper. Beyond reproducing the reported experiments, the released evaluation pipeline and protocol documentation are intended to let practitioners apply the three-axis protocol to their own supervised reconstruction models, on benchmarks beyond the two studied here.

\subsection{Code repository and structure}
\label{app:code}
%The code accompanying this paper is released at \href{https://anonymous.4open.science/r/evaluating-generative-models-C7C8/}{anonymous.4open.science/r/evaluating-generative-models-C7C8/} (MIT License).\footnote{For double-blind review, hosted via \href{https://anonymous.4open.science}{anonymous.4open.science}; the public URL will be substituted on acceptance.}
The code accompanying this paper is available from the corresponding author on request.
The repository contains the implementations of all learned models in Benchmark~II, the training and evaluation pipelines, the reference implementations of the three protocol metrics, and the analysis notebooks used to produce every table and figure in \cref{sec:toy-model,sec:top-reconstruction}. Compute requirements, training times, and hyperparameters are summarised in \cref{app:benchmark-1-model-architectures-and-training-details,app:benchmark-2-training-details}. Executed notebooks and the pre-computed evaluation arrays underlying every Benchmark~II table and figure are available from the corresponding author on request.
%on Zenodo\footnote{For double-blind review, we provide a tokenized Zenodo link. DOI will be substituted on acceptance.} under CC-BY 4.0, enabling readers to regenerate the plots and verify reported metrics without rerunning training or posterior sampling:
%\url{https://zenodo.org/records/20044010?token=eyJhbGciOiJIUzUxMiJ9.eyJpZCI6IjExZTBiNDc0LTdlMWEtNDhhNS04MTJiLWNlYWRlOTAwN2UyNyIsImRhdGEiOnt9LCJyYW5kb20iOiI4YzliMWNlMWEzODQ2OGFkNzk1NmJkMzkyZTlkYzk5ZSJ9.Sn5opswN_ujCrJ1gqDhHlhzKhPjA_Iiak2TdSJpo6nPIpARlfC6yGsZO8Pgc3xnJY-4jJEtqNRyFPmkjzfVSNg}

\paragraph{Repository layout}
\begin{itemize}
    \item \texttt{README.md} A more practical oriented description on how the repository is set up, how to install the virtual environment, and how to run different parts of the code base.
    \item \texttt{src/top\_reco/} -- the core Python package for the studies presented in Benchmark~II. The \texttt{models/} subpackage contains the model implementations; \texttt{losses/}, \texttt{tasks/} and \texttt{data/} provide the training objectives, the dataset wiring and the Delphes-format data loader. 
    
    The \texttt{metrics/} subpackage provides reference implementations of CRPS (\cref{sec:crps}), spectrum fidelity (\cref{sec:spectrum-fidelity}), and the conformal calibration diagnostic (\cref{sec:coverage}), each independently testable and importable as \texttt{from top\_reco.metrics import \dots}.
    \item \texttt{configs/} -- The model configuration tree with hyperparameters for each model.
    \item \texttt{scripts/} -- \texttt{train.py} (entry point for training a model), \texttt{evaluate.py} (produces the per-event metric arrays), \texttt{posterior\_sampling.py} (produces the dense posterior samples used for the per-event posterior figures).
    \item \texttt{notebooks/} -- \texttt{benchmark1\_synthetic.ipynb} is end-to-end and self-contained (no external data). It produces all tables and figures related to Benchmark~I. \texttt{benchmark2\_plotting.ipynb} reads pre-computed evaluation arrays from \texttt{data/benchmark2/} and produces all tables and figures related to Benchmark~II.
    \item \texttt{tests/} -- unit tests for the \texttt{coordinates/} and \texttt{metrics/} modules.
    \item \texttt{docs/} -- Sphinx-built API reference for the metrics module, generated from the same docstrings that document the formulae of \cref{sec:three-step-evaluation-protocol}.
    \item \texttt{NOTICE.md} and \texttt{LICENSES/} -- third-party code attribution; portions of \texttt{src/top\_reco/coordinates/} and the transformer building blocks of \texttt{src/top\_reco/models/} are adapted from other sources.
\end{itemize}

The \emph{analytic solver} baseline used in Benchmark~II is the algebraic on-shell-mass solver of \citep{sonnenschein-PhysRevD.72.095020}; %the implementation is external to this repository and is available at \href{https://anonymous.4open.science/r/analyticsolver-607E/}{anonymous.4open.science/r/analyticsolver-607E} (MIT License).\footnote{For double-blind review, hosted via \href{https://anonymous.4open.science}{anonymous.4open.science}; the public URL will be substituted on acceptance.}
the implementation is external to this repository and is available from the corresponding author on request.

\subsection{Compute resources}
All experiments were run on a single \texttt{NVIDIA A100 80GB PCIe} GPU. Per-model training times are: Benchmark~I -- MLP-backbone models ~1 minute each, normalizing flow ~7 minutes (5 seeds each); Benchmark II -- transformer regressions ~4 hours, continuous flow ~16 hours, discrete flow ~8 hours (single seed each, table 6). In addition, the flow-based models needed GPU-time for sampling which we estimate to a combined sampling time of 30 GPU-hours. Total reported-experiment compute is approximately 65 GPU-hours. Total project compute including hyperparameter exploration, MMD-bandwidth scans on additional observables (\cref{app:benchmark-2-model-architectures}), and discarded architectural variants is approximately 400 GPU-hours.

\subsection{Dependencies and environment}
\label{app:dependencies-and-environment}
Models are implemented in PyTorch ($\geq$\texttt{2.2}). The synthetic-benchmark normalizing flow uses normflows \citep{Stimper2023}. The continuous flow uses standard PyTorch ODE integration utilities \citep{torchdiffeq}. Full dependency lists with pinned versions are provided in the repository in \texttt{uv.lock}.

\subsection{Dataset licensing and access}
The Benchmark~II dataset is the public Delphes Monte Carlo release of \citet{delphes-dataset}, distributed under CC-BY~4.0. We use the train/test splits provided with the release without modification, and we restrict ourselves to events generated with MadGraph. The Benchmark~I synthetic dataset is generated by the released code from the forward model of \cref{eq:benchmark-1-forward-model}; the data-generation script is included in the repository and produces deterministic outputs given the documented seeds.

%%%%%%%%%%%%%%%%%%%%%%%%%%%%%%%%%%%%%%%%%%%%%%%%%%%%%%%%%%%%

%%%%%%%%%%%%%% 4open/zenodo/checklist (searchqueries to remember) %%%%%%%%%%%%%%
%%%%%%%%%%%%%% TODO: Uncomment in camera-ready %%%%%%%%%%%%%%
%\newpage
%\input{checklist.tex}

\end{document}